\documentclass{article}

\PassOptionsToPackage{table,dvipsnames}{xcolor}
\usepackage{style}

\usepackage[utf8]{inputenc}
\usepackage[T1]{fontenc}
\usepackage{url}
\usepackage{booktabs}
\usepackage{amsfonts}
\usepackage{nicefrac}
\usepackage{microtype}
\usepackage[table,dvipsnames]{xcolor}

\usepackage{mdframed}
\usepackage{graphicx}
\usepackage{subcaption}
\usepackage{enumitem}
\usepackage{amsmath}
\usepackage{pifont}
\usepackage{makecell}
\usepackage{tikz}
\usepackage{simpleicons}
\usepackage{fontawesome5}
\usetikzlibrary{arrows.meta,positioning,fit}

\newcommand{\done}{\textcolor{ForestGreen}{\ding{51}}}
\newcommand{\fail}{\textcolor{red}{\ding{55}}}
\newcommand{\githuburl}{https://github.com/dhevarghese/PlaySuite}
\newcommand{\hfurl}{https://huggingface.co/datasets/Dheerraa/PlaySuite}
\newcommand{\playsuite}{\textsc{PlaySuite }}

\usepackage[colorlinks=true,
    linkcolor=accent,
    citecolor=accent,
    filecolor=accent,
    urlcolor=accent]{hyperref}
\hypersetup{pdftitle={PlaySuite: A Large-Scale Benchmark for Interactive Visual Intelligence},
  pdfauthor={Dheeraj Varghese, Anna Vettoruzzo, Walter Simoncini, Michelle Lorena Acevedo Callejas, Mohammad Mahdi Derakhshani, Kristof Meding, Joaquin Vanschoren, Cees G. M. Snoek}}
\usepackage{placeins}

\usepackage{needspace}
\usepackage{listings}

\lstdefinestyle{promptjson}{basicstyle=\ttfamily\footnotesize,
  frame=none,backgroundcolor={},numbers=none,
  xleftmargin=0pt,xrightmargin=0pt,aboveskip=3pt,belowskip=3pt,
  breaklines=true,breakatwhitespace=true,columns=fullflexible,
  keepspaces=true,showstringspaces=false}

\begin{document}

\begin{wtctitleblock}
\wtctitle{\smash{\raisebox{-0.22\height}{\includegraphics[height=1.35em]{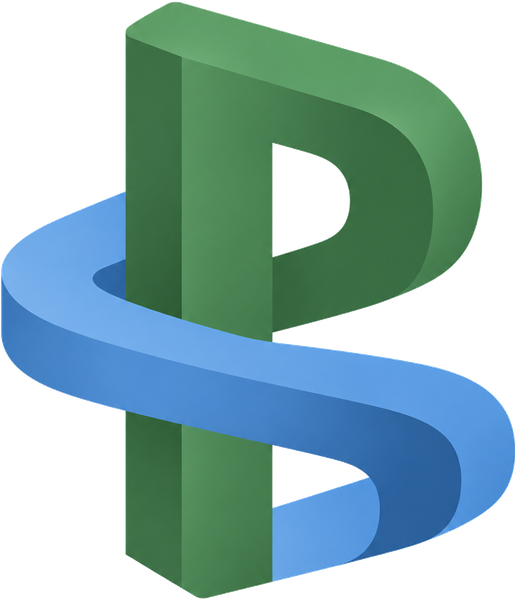}}}\hspace{0.25em}\textcolor[HTML]{5D8D61}{Play}\textcolor[HTML]{4275C5}{Suite}}{A Large-Scale Benchmark for Interactive Visual Intelligence}
\begin{center}
{\sffamily\bfseries
  \mbox{Dheeraj Varghese\textsuperscript{1}},\quad
  \mbox{Anna Vettoruzzo\textsuperscript{2,*}},\quad
  \mbox{Walter Simoncini\textsuperscript{1,*}},\quad
  \mbox{Michelle Lorena Acevedo Callejas\textsuperscript{3}},\quad
  \mbox{Mohammad Mahdi Derakhshani\textsuperscript{1}},\quad
  \mbox{Kristof Meding\textsuperscript{3,$\dagger$}},\quad
  \mbox{Joaquin Vanschoren\textsuperscript{2,$\dagger$}},\quad
  \mbox{Cees G.~M.~Snoek\textsuperscript{1,$\dagger$}}\par}
\smallskip
{\sffamily\small
  \textsuperscript{1}University of Amsterdam\quad
  \textsuperscript{2}Eindhoven University of Technology\quad
  \textsuperscript{3}Eberhard Karls Universit\"at T\"ubingen\par}
\end{center}
\begin{abstract}
Recent advances in multimodal foundation models yield strong performance on static perception and reasoning benchmarks, yet such evaluations largely overlook a central aspect of intelligence: acting competently in dynamic environments over extended time horizons. We introduce \textsc{PlaySuite}, a large-scale benchmark for evaluating interactive visual intelligence across more than 5K open-source video games curated from \emph{PyWeek} and \emph{itch.io}. Spanning diverse genres and engines, including Pygame, HTML5, Godot, and Unity, these independent games are largely out-of-distribution for current models, reducing the likelihood that success can be achieved by retrieving memorized walkthroughs or web-scale training artifacts. To enable scalable evaluation across heterogeneous titles, we develop a unified closed-loop interaction framework optimized for HPC clusters alongside a Video-LLM-as-a-judge protocol that maps observable gameplay milestones to standardized progress levels. We evaluate fourteen recent open models spanning vision-language models, computer-use agents, and vision-language-action models. Our results yield strong evidence of a perception-action gap: despite strong reasoning capabilities, current models struggle to make sustained progress and exhibit recurring failures in spatial grounding, action execution, and self-correction. \playsuite  provides a reproducible and extensible testbed for measuring progress from visual perception to goal-directed interaction, and a foundation for developing models that can act, adapt, and generalize in dynamic visual environments.
\end{abstract}
\wtctitlefooter{%
  \textsuperscript{*}Equal contribution (shared second authors).\quad
  \textsuperscript{$\dagger$}Equal advising (shared last authors).\\[4pt]
  {\fontsize{9}{11}\selectfont
  \href{\githuburl}{{\color{accent}\simpleicon{github}}~\textbf{GitHub}}\qquad
  \href{\hfurl}{{\color{accent}\simpleicon{huggingface}}~\textbf{Hugging Face}}\qquad
  \href{mailto:d.varghese@uva.nl}{{\color{accent}\faEnvelope[regular]}~\textbf{d.varghese@uva.nl}}}
}
\end{wtctitleblock}
\vspace{-4pt}
\begingroup
\noindent\includegraphics[width=\textwidth]{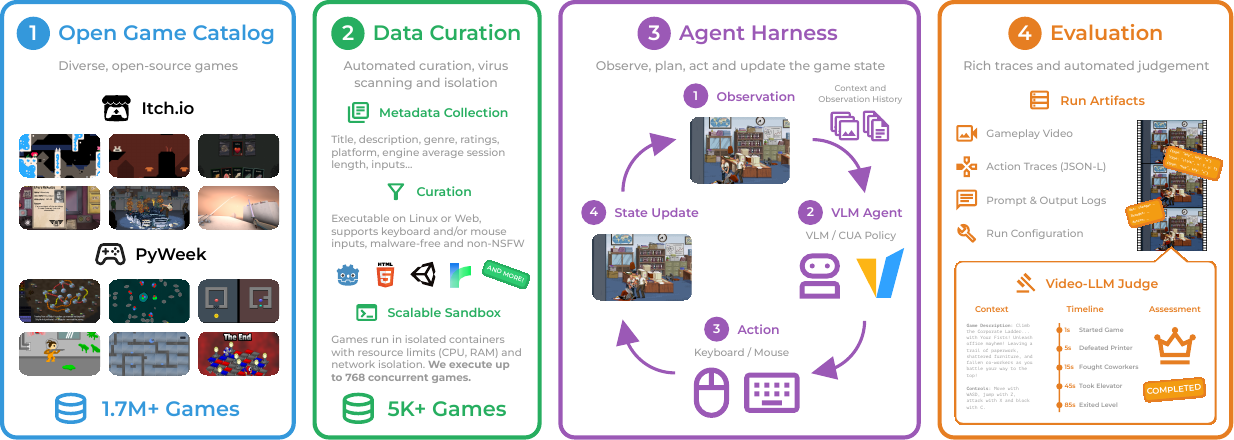}
\captionof{figure}{\textbf{The \playsuite benchmark.} Our open-source benchmark for interactive visual intelligence comprises four stages:
(1) Open Game Catalog: A diverse library of 1.7M video games sourced from \emph{itch.io} and \emph{PyWeek}.
(2) Data Curation: An automated filtering and sandboxing pipeline that distills the catalog into approximately 17.5K executable, safe, and reproducible environments, of which 5,734 are used in our evaluation.
(3) Interactive Agent Harness: A high-performance, closed-loop system where foundation models interact via standard keyboard/mouse interfaces for optimized large-scale HPC evaluation.
(4) Evaluation Artifacts: A comprehensive suite of gameplay videos, action traces, and a Video-LLM-as-a-judge mechanism to quantify goal-directed performance.}
\label{fig:playsuite}
\par\endgroup

\section{Introduction}

Recent advances in multimodal foundation models have rapidly expanded their capabilities beyond static visual understanding toward multimodal reasoning and interaction. Modern models can operate graphical user interfaces across desktop, mobile, and web platforms and control embodied agents over extended tasks~\citep{bai2025qwen2,zhou2026qwen, qin2025ui,intelligence2025pi_}. Yet, many multimodal benchmarks still primarily evaluate perception and reasoning over static inputs \citep{yue2024mmmu, liu2024mmbench}. Real-world intelligence~\citep{lake2017building}, by contrast, is interactive and temporally dependent: actions alter future observations, decisions require adaptation to feedback, and agents must recover from mistakes over time. Recent agent benchmarks~\citep{xie2024osworld,zhouwebarena,zheng2025v,zhang2025videogamebench} are increasingly incorporating assessment of such closed-loop capabilities, but doing so across large and diverse sets of environments remains challenging. This motivates a central question: \textit{how well do current multimodal models translate their perceptual and reasoning capabilities into goal-directed interaction?}

Games provide a particularly rich testbed to address this question. They combine computer use with spatial reasoning, temporal dynamics, and goal-directed behavior in interactive environments. Recent frontier systems have demonstrated sustained interaction in complex games from visual observations and natural-language goals~\citep{tan2024cradle,bolton2025sima,anthropic2026fable, openai2026astra}, but these demonstrations remain concentrated in a small number of environments. Existing game benchmarks similarly span only a limited number of distinct games~\citep{Guss_Houghton_Topin_Wang_Codel_Veloso_Salakhutdinov_2019,fan2022minedojo,zheng2025v,paglieribalrog,park2025orak,zhang2025videogamebench,wang2026visgym}, raising questions on how broadly these capabilities generalize. We introduce \textsc{PlaySuite}, a large-scale benchmark for interactive visual intelligence across 5,734 distinct video games, over two orders of magnitude more game environments ($> 100\text{x}$) than prior multi-game benchmarks. These environments are drawn from a larger pool of approximately 17.5K launchable games validated by our pipeline, providing substantial room for future expansion. An illustration of \playsuite is presented in Figure~\ref{fig:playsuite}.

We construct \playsuite from games hosted on~\cite{pyweek} and~\cite{itchio}. PyWeek contributes lightweight Python-based games, while itch.io substantially expands the visual, mechanical, and engine diversity of the benchmark with games developed using Unity, Godot, GameMaker, RPG Maker, and other frameworks. Together, these ecosystems provide a broad distribution of independently developed, human-designed environments, ranging from simple 2D games to visually complex 3D and physics-driven settings. Evaluating agents across thousands of heterogeneous games introduces two key challenges: scalable interaction and comparable evaluation. Different engines expose disparate execution and control interfaces, motivating a common interaction layer that enables agents to operate across games using standard keyboard and mouse actions. At the same time, games do not share a common reward or scoring interface. We therefore introduce a progress-based Video-LLM-as-a-judge methodology~\citep{zheng2023judging,lin2023mcu}, which extracts observable gameplay milestones and maps them to progress levels. Together, these components enable scalable and comparable evaluation across thousands of diverse game environments.

Using this new infrastructure, we benchmark fourteen recent open models spanning general-purpose vision-language models (VLMs), computer-use agents (CUAs), and vision-language-action models (VLAs). Despite strong visual reasoning capabilities, current models achieve meaningful gameplay progress on only a minority of games. Qualitative inspection of gameplay traces further identifies recurring difficulties in spatial grounding, control selection, action execution, and adaptation over time, highlighting the gap between interpreting an environment and acting effectively within it.

Nearly a decade after OpenAI Universe~\citep{openai2016universe}, \playsuite revisits the vision of broad-environment evaluation in the setting of pretrained multimodal agents that can be deployed across heterogeneous games without environment-specific training. We release \playsuite as an open and extensible benchmark, including environment wrappers, evaluation protocols, and high-performance cluster tooling. In summary, our contributions are threefold: (1) a large-scale benchmark spanning over 5K heterogeneous interactive visual environments; (2) a unified, HPC-optimized infrastructure for closed-loop evaluation across diverse game engines; and (3) a scalable milestone-based Video-LLM evaluation protocol for measuring gameplay progress without environment-specific rewards. The benchmark is designed to grow as new environments and increasingly capable interactive models emerge.

\section{Related Work}

\noindent\textbf{Foundation models in interactive environments.}
LLMs have shifted from passive information retrieval toward agentic interaction in closed-loop systems~\citep{wang2024survey}. Early work focused on reasoning-chain planning, decomposing natural language goals into executable sub-steps~\citep{huang2022language, xie2023translating}, but requiring recursive feedback to adapt to  environment shifts~\citep{yao2023react, wang2023describe}. With the emergence of multimodal foundation models, research has shifted toward unified perception-action policies \citep{mokaria2026survey}. General VLMs~\citep{bai2025qwen2, gemma2025} ground linguistic instructions in raw visual pixels, while specialized GUI agents~\citep{qin2025ui,lin2025showui, wuatlas} predict control tokens directly. Despite these advancements, models still struggle with the fundamental challenges of interactive intelligence: translating granular perception into effective action, maintaining long-horizon planning, and exhibiting robust spatial reasoning in dynamic environments. This persistent gap highlights the need for an open-source, large-scale, and heterogeneous benchmark to evaluate and refine these capabilities across interactive scenarios, moving beyond narrowly defined domains toward truly general-purpose agency.

\noindent\textbf{Game-based benchmarks.}
To address performance saturation and data contamination in traditional evaluations, the field has pivoted toward dynamic, interactive suites that require real-time decision-making~\citep{wong2025reasoning, chen2025recent}. Digital navigation benchmarks~\citep{he2024webvoyager, zhang2025appagent, zhouwebarena, yao2022webshop} assess tool-use and GUI automation, while embodied AI frameworks focus on atomic physical interactions~\citep{yang2025embodiedbench}. Games have emerged as a primary testbed due to their inherent complexity and high-frequency temporal dynamics. Early benchmarks were restricted to 2D or grid-based environments~\citep{costarelli2024gamebench, nasir2024gametraversalbenchmark, hafner2021benchmarking, kuttler2020nethack, wang2025large, chollet2025arc}. While these provided essential testbeds for logic and planning, they failed to capture the spatial nuances of the real world~\citep{zhang2024ing, wang2025large}. Recent 3D environments require agents to process perspective, depth, and occlusion~\citep{qiao2023gameeval, hugamearena, Xie_Zhang_Chen_Wan_Li_2024}, but remain constrained to single titles~\citep{fan2022minedojo, Guss_Houghton_Topin_Wang_Codel_Veloso_Salakhutdinov_2019}, or small sets of games requiring bespoke per-title engineering and hand-crafted rewards~\citep{zheng2025v, paglieribalrog, park2025orak, ying2026ai}.
Modern modular harnesses decouple perception and reasoning to tackle diverse games without domain-specific engineering~\citep{zhang2025general, zhang2025videogamebench, lu2025cultivating, foundation2026arc}, often adopting the LLM/VLMs-as-a-judge paradigm to assess agentic competence~\citep{zheng2023judging,pan2024autonomous,lin2023mcu}. \playsuite synthesizes these advancements into a scalable paradigm, replacing bespoke engineering with a unified, judge-based assessment across a large variety of game mechanics. A comprehensive comparison of the existing video games benchmarks against \playsuite is provided in Table~\ref{tab:literature_comparison}, highlighting our contributions in scale, infrastructure, and dynamic evaluation.

\begin{table}[t]
\centering
\small
\setlength{\tabcolsep}{4.2pt}
\renewcommand{\arraystretch}{1.12}
\caption{\textbf{Related work.} Comparison of \playsuite against existing video game benchmarks.
}
\label{tab:literature_comparison}
\resizebox{\linewidth}{!}{
\begin{tabular}{lcrccccccc}
\toprule
& & & \multicolumn{4}{c}{\textbf{Models}} & \multicolumn{3}{c}{\textbf{Evaluation}}\\
\cmidrule(lr){4-7} \cmidrule(lr){8-10}
\textbf{Benchmark}
& \textbf{Domain}
& \textbf{\# Games}
& LLM
& VLM
& VLA
& CUA
& Metric
& HPC
& Dynamic \\
\midrule
GameBench~\citep{costarelli2024gamebench}
& Text & 9 & \done & \fail & \fail & \fail & Game score & \fail & \fail \\

GameArena~\citep{hugamearena}
& Text & 6 & \done & \fail & \fail & \fail & Win rate & \fail & \fail \\

Balrog~\citep{paglieribalrog}
& 2D & 6 & \done & \done & \fail & \fail & Task completion & \fail & \fail \\

ING-VP~\citep{zhang2024ing}
& 2D & 6 & \fail & \done & \fail & \fail & Task completion & \fail & \fail \\

V-MAGE~\citep{zheng2025v}
& 2D/3D & 5 & \fail & \done & \fail & \fail & ELO ranking & \fail & \fail \\

VS-Bench~\citep{xu2025vs}
& 2D/3D & 10 & \fail & \done & \fail & \fail & Game score & \fail & \fail \\

Orak~\citep{park2025orak}
& 2D/3D & 12 & \done & \done & \fail & \fail & Game score & \fail & \fail \\

VisGym~\citep{wang2026visgym}
& 2D/3D & 17 & \fail & \done & \fail & \fail & Success rate & \fail & \fail \\

VideoGameBench~\citep{zhang2025videogamebench}
& 2D/3D & 23 & \fail & \done & \fail & \fail & Progress tracking & \fail & \done \\
\midrule
\rowcolor{gray!8}
\playsuite
& 2D/3D
& \textbf{5,734}
& \fail & \done & \done & \done
& Automated judge
& \done
& \done \\
\bottomrule
\end{tabular}
}
\end{table}

\section{\playsuite Construction}
\label{sec:playsuite_bench}

\begin{figure}[t]
\centering
\includegraphics[width=\textwidth]{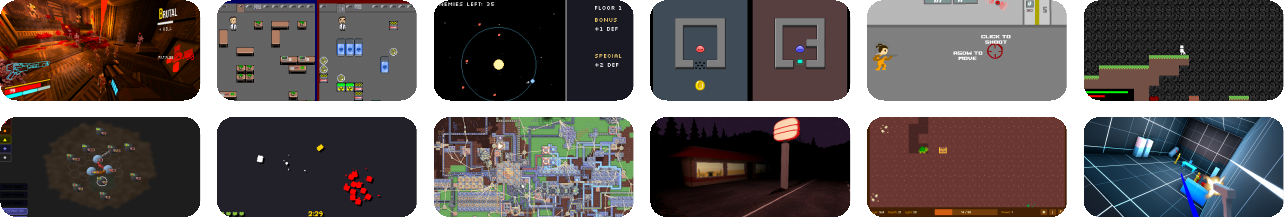}
\caption{\textbf{Examples from the \playsuite game corpus.}
\playsuite contains a broad collection of independently developed games with varied visual styles, mechanics, camera perspectives, and controls. These examples illustrate the range of settings agents must handle, from 2D platform and puzzle games to racing, exploration, shooting, and 3D navigation environments. }
\label{fig:games-collection}

\end{figure}

\playsuite is a large-scale benchmark designed to evaluate general-purpose interactive agents across a vast distribution of game environments. Unlike existing benchmarks that rely on domain-specific simulators or restricted game lists, \playsuite provides a unified interaction framework that standardizes communication across heterogeneous engines. Each game is executed in an isolated sandbox using a virtual display, enabling efficient and scalable execution across GPU clusters without requiring physical displays. To ensure the benchmark captures a broad spectrum of common-sense interactive intelligence across different tasks, we curated a library of 5,734 open-source games from ~\cite{pyweek} and ~\cite{itchio}. These platforms offer a unique diversity of 2D and 3D dynamics that are designed for human players to learn intuitively within seconds, making them ideal for testing the zero-shot performance of multimodal foundation models, see examples in Figure~\ref{fig:games-collection}.

\noindent\textbf{Filtering and quality control.}
From an initial crawl, we selected games with a community rating $>3.0 / 5.0$ and at least $5$ ratings to ensure playability and coherent goal design. We filtered for platform compatibility, retaining only Linux and Web-based games, and strictly excluded NSFW content and games with broken assets through a multi-stage automated and manual verification process. To ensure secure execution at scale, we implemented a custom safety triage pipeline that assigns a risk score to each game based on structural and metadata signals. High-risk archives were quarantined, while all others were executed in a network-isolated sandbox; see Appendix~\ref{app:data_curation}.

\noindent\textbf{Taxonomy and categorization.} The collection spans nine distinct genres: \texttt{Action, Adventure, Puzzle, Shooter, Simulation, Survival, Strategy, Fighting, Racing}. Our taxonomy was derived from a cross-analysis of prior work on game-based AI~\citep{Lee2014facet, hu2024survey} and frequent tags used by the developer community. This ensures that the benchmark covers diverse forms of interaction and decision-making. For games from \emph{itch.io}, we used existing metadata tags. For \emph{PyWeek} games, which lack native tagging, we employed a frontier LLM to categorize each title based on its source code documentation and README files. Further details are provided in Appendix~\ref{app:taxonomy_categories}. This diverse distribution ensures that models cannot succeed through narrow heuristics tailored to a single genre. Genre counts are shown in Appendix Figure~\ref{fig:game_categories_comparison}.

\noindent\textbf{Evaluation pipeline.}
We formalize the evaluation as a sequential decision-making process in which a model functions as an agent interacting with a game environment $\mathcal{E}$. At each time step $t \in [0, T]$, the model receives an input $\mathcal{I}_t = \langle p, o_t, h_t \rangle$, where $p$ is a unified prompt template with game-specific metadata, $o_t \in \mathbb{R}^{H \times W \times 3}$ is the raw pixel frame, and  $h_t=\{(o_{t-k}, a_{t-k}), \dots, (o_{t-1}, a_{t-1})\}$ is an optional temporal history of previous observations and actions. Crucially, models are not provided with solution strategies; they must autonomously infer mechanics and objectives from visual feedback and metadata.
The model generates a structured output $\mathcal{O}_t = \langle r_t, a_t \rangle$, consisting of a reasoning trace $r_t$ and an action $a_t$. This defines a policy $\pi(\mathcal{O}_t \mid p, o_t, h_t)$ that must generalize across the transition dynamics of thousands of environments. By requiring the agent to bridge the gap between seeing the current state $o_t$ and doing the correct action $a_t$ without bespoke tuning, this protocol shifts the focus from static recognition to functional, interactive intelligence under varied control dynamics.

\noindent\textbf{Milestone-based evaluation.}
\label{sec:harness_judge}
A key challenge in evaluating model performance across thousands of heterogeneous environments is the lack of a standardized reward signal. We address this with a scalable Video-LLM-as-a-judge framework that measures progress from observable gameplay events. The judge reports timestamped evidence of events such as score increases, entry into a new area, checkpoint or objective completion, and game completion. A fixed rule then maps these observations to five ordered progress levels: \texttt{None} (0), \texttt{Minimal} (1), \texttt{Partial} (2), \texttt{Substantial} (3), and \texttt{Completed} (4). Partial progress requires either reaching a new area or level, or increasing the score; Substantial progress requires completing an intermediate objective or observing both forms of progress; and Completed requires explicit visual evidence of game completion. Minimal progress is assigned when player control is evident through at least two distinct interactions.

Our evaluation setup uses Qwen3.5-9B with the milestone prompt, structured JSON decoding, and video sampled at 1\,fps. We report the mean Progress Score across valid evaluations and the fraction of runs reaching at least Partial progress. The mean Progress Score summarizes ordinal outcomes across the benchmark and should not be interpreted as the fraction of a game completed. The complete prompt, schema, and scoring implementation are provided in Appendix~\ref{app:judge}.
To complement automated evaluation, we provide an inspection toolkit that overlays gameplay videos with the agent's generated responses and actions together with the judge's timestamped assessments. This allows us to audit evaluations, inspect model and judge behavior, and provide human feedback or corrections starting from the automated analysis rather than reviewing trajectories from scratch.

\noindent\textbf{Judge validation.}
We validate Qwen3.5-9B against six human annotators on 60 gameplay videos, comprising 45 \emph{itch.io} and 15 \emph{PyWeek} runs spanning 14 model presets and both history conditions. Human-only Krippendorff's $\alpha$ is 0.624, while including the judge as a seventh rater yields $\alpha=0.605$. In pairwise comparisons, the judge's progress rating is within one level of the human rating in 90.4\% of cases, compared with 93.0\% for human--human pairs. We additionally evaluate Qwen3.5-27B, Gemma-4-31B-it, and Molmo2-8B. Qwen3.5-27B achieves the highest agreement with humans at 91.6\% within one level, compared with 90.4\% for Qwen3.5-9B. We therefore use Qwen3.5-9B as the default judge, providing comparable agreement while remaining more practical for evaluation at scale. See Appendix~\ref{app:judge_validation} for full results.

\section{\playsuite Experimental Architecture}
\label{sec:ExperimentalArchitecture}

To maintain a scalable and engine-agnostic evaluation across the environments, we developed a standardized experimental architecture. This setup ensures that model success is a product of general interactive intelligence rather than per-game optimization.

\noindent\textbf{Input and output configuration.}
For models evaluated through the standard prompt, we use shared system and user templates that insert the game title, taxonomic genre, available developer-provided metadata or instructions, the current visual observation, and optional textual history of previous interactions. Unlike benchmarks that provide predefined game-specific control mappings~\citep{zheng2025v,park2025orak,zhang2025videogamebench}, \playsuite does not assume prior knowledge of a game's mechanics. When developer documentation specifies the controls, the prompt instructs the model to follow them. When controls are absent, the model must infer them through visual context and exploration, aided by fallback keyboard controls and a set of decision heuristics for choosing between keyboard and mouse actions. The prompt also defines the supported action grammar and instructs the model to change strategy when previous actions produce no visible progress. This setup approximates how a player approaches an unfamiliar game with limited documentation while ensuring that the model's actions remain executable by the evaluation harness.

The action grammar includes key presses, simultaneous key combinations, timed holds, absolute and relative mouse movement, left/right clicks, scrolling, and repeated clicks. Dragging is unsupported, so we exclude games whose controls mention dragging. The standard template requires to generate a structured response with Observation, Reasoning, and Action fields. This includes a reasoning trace, which incorporates both a high-level summary of the current observation and the underlying logic for the subsequent move, followed by the next predicted action. The harness parses the action for execution and retains the complete raw response, including the observation and reasoning fields when present. Models with native computer-use or visual-action interfaces retain their model-specific prompts, output formats, and coordinate conventions. Model-specific adapters translate their predicted actions into the shared keyboard and mouse interface. More details about the prompt templates, model assignments, and parsers can be found in Appendix~\ref{app:prompts}.

\noindent\textbf{Temporal context and history.}
To facilitate temporal reasoning and long-horizon planning, we incorporate a rolling history of previous interactions, $h_t = \{(o_{t-k}, a_{t-k}), \dots, (o_{t-1}, a_{t-1})\}$. This history is intended to help the model capture causal relationships between actions and environmental transitions. We empirically evaluated history lengths of $k \in \{0, 3, 5, 10\}$ to determine the optimal balance between temporal awareness and context window efficiency.

\noindent\textbf{Baseline models.}
We benchmark 14 models spanning three broad families of interactive vision models: general-purpose VLMs, CUAs, and VLAs. Our evaluation includes VLMs such as \emph{Gemma-4-26B-A4B-it}, \emph{Gemma-4-E4B-it}, \emph{Qwen3-Omni-30B-A3B-Instruct}, \emph{Qwen3-VL-8B-Instruct}, \emph{Qwen2.5-VL-7B-Instruct}, \emph{Gemma-3-12B-it}, \emph{Qwen2.5-Omni-7B}, and \emph{InternVL2.5-8B}, alongside models explicitly designed for computer interaction, including \emph{EvoCUA-8B}, \emph{UI-TARS-1.5-7B}, \emph{OpenCUA-7B}, \emph{GUI-Owl-1.5-8B-Instruct}, \emph{OS-Atlas-Pro-7B}, and \emph{ShowUI-2B}. All models are run with a temperature $0.5$ and a maximum output length of $512$ tokens. By including these varied architectures, we provide a broad comparison of how different modeling priors handle the zero-shot challenges of interactive game environments. We restrict evaluation to open-weight models, thereby excluding proprietary systems such as Cradle~\citep{tan2024cradle} and SIMA 2~\citep{bolton2025sima}. Appendix~\ref{app:models} lists the exact model identifiers, prompt templates, and adapters.

\noindent\textbf{Execution and interface setup.}
The evaluation pipeline is designed for high-throughput, display-independent execution on HPC infrastructure. Across both subsets, game progression is suspended during model inference. This decouples game time from model-serving latency and ensures that each selected action is applied to the state observed by the model. To accommodate heterogeneous game engines, the pipeline supports two execution formats, detailed in Appendix~\ref{app:execution}. In both, model-specific adapters map generated actions to a shared action representation, after which a unified virtual input driver emits either low-level \texttt{pygame} events or system-level keyboard and mouse events.

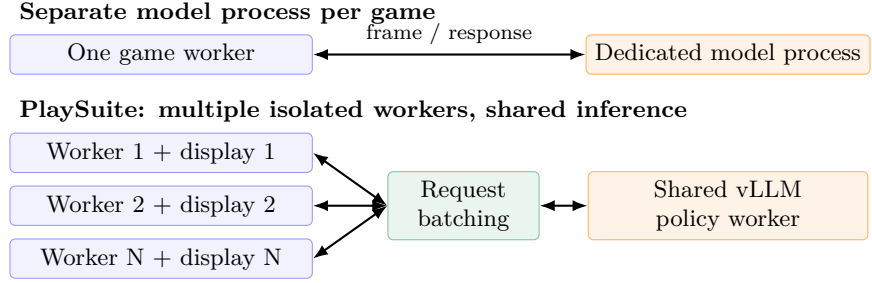
\begin{figure}[t]
\centering
\begin{tikzpicture}[
  font=\small,
  nodebox/.style={
    draw,
    rounded corners=2pt,
    align=center,
    minimum height=.52cm
  },
  worker/.style={
    nodebox,
    fill=blue!5,
    draw=blue!40,
    minimum width=4.0cm
  },
  inference/.style={
    nodebox,
    fill=orange!10,
    draw=orange!60,
    minimum width=3.7cm
  },
  arrow/.style={-{Latex[length=2mm]},thick},
  sync/.style={{Latex[length=2mm]}-{Latex[length=2mm]},thick}
]
\node[font=\bfseries\small,anchor=west]
  at (-2.0,2.2) {Separate model process per game};
\node[worker] (single)
  at (0,1.65) {One game worker};
\node[inference] (dedicated)
  at (7.5,1.65) {Dedicated model process};
\draw[sync]
  (single) --
  node[above,font=\footnotesize]{frame / response}
  (dedicated);
\node[font=\bfseries\small,anchor=west]
  at (-2.0,.9) {PlaySuite: multiple isolated workers, shared inference};
\node[worker] (w1)
  at (0,.35) {Worker 1 + display 1};
\node[worker] (w2)
  at (0,-.35) {Worker 2 + display 2};
\node[worker] (wn)
  at (0,-1.05) {Worker N + display N};
\node[
  nodebox,
  fill=ForestGreen!7,
  draw=ForestGreen!50,
  minimum width=2.0cm
] (queue)
  at (4.0,-.35) {Request\\batching};
\node[inference] (shared)
  at (7.5,-.35) {Shared vLLM\\policy worker};
\foreach \w in {w1,w2,wn}{
  \draw[sync] (\w.east) -- (queue.west);
}
\draw[sync] (queue) -- (shared);
\end{tikzpicture}%
\caption{\textbf{Batched multi-worker execution.} Comparison between a sequential one-game-per-model-process evaluation pattern and the \playsuite batched architecture. Multiple sandboxed game workers run on one node, each with an isolated virtual display, while a shared vLLM-backed inference worker batches ready frame-and-prompt requests on the GPU.}

\label{fig:hpc_architecture}
\end{figure}

\noindent\textbf{Batched multi-worker execution.}
Figure~\ref{fig:hpc_architecture} illustrates how \playsuite parallelizes closed-loop evaluation on HPC infrastructure. A naive closed-loop evaluation launches one game and one model process per node, duplicating model weights, under-utilizing the GPU during game rendering and environment startup. In contrast, we launch many independent game workers on a single HPC node. Each worker is assigned a unique virtual display, an isolated output directory, and a per-worker sandbox scratch space. Workers capture frames, construct model prompts, inject parsed keyboard or mouse actions, and write the resulting video and action trace.

Model inference is managed by a shared vLLM worker. Lightweight game workers send frame-and-prompt requests to this server over a Unix socket so that game processes can remain network-isolated. The inference worker holds a single copy of the model in GPU memory, batches ready requests, and returns raw model outputs to the workers, where model-specific parsers translate text into the common action interface. This design amortizes model loading across games, improves GPU utilization, and separates CPU-bound game execution from GPU-bound inference. In a controlled evaluation with Qwen3-VL-8B on one H100, eight concurrent workers increase throughput from 6.7 to 42.4 evaluations per GPU-hour. Additional measurements of launch, gameplay, and judging throughput are provided in Appendix~\ref{app:execution_analyses}.

\section{\playsuite Benchmark Results}
\label{sec:results}

We evaluate fourteen open models on \playsuite across 5,630 \emph{itch.io} games and 104 \emph{PyWeek} games. We report the mean Progress Score on the five-level scale from \emph{None} (0) to \emph{Completed} (4), together with the percentage of runs reaching at least \emph{Partial} progress.

\textbf{Overall performance.}
\begin{table*}[t]
    \centering
    \small
    \caption{
        \textbf{Main benchmark results across \playsuite.}
        Mean Progress Score under the history ($k=3$) and no-history
        conditions, together with the percentage of no-history runs reaching at least \emph{Partial} progress, across 5,630 \emph{itch.io} games and 104 \emph{PyWeek} games. Progress is measured on the five-level ordinal scale from \emph{None} (0) to \emph{Completed} (4). Best results are shown in bold.
    }
    \label{tab:overall_results}

    \setlength{\tabcolsep}{4.5pt}
    \renewcommand{\arraystretch}{1.08}

    \begin{tabular}{@{}lrrrrrr@{}}
        \toprule
        & \multicolumn{3}{c}{\textbf{\emph{itch.io}}}
        & \multicolumn{3}{c}{\textbf{\emph{PyWeek}}} \\
        \cmidrule(lr){2-4}
        \cmidrule(lr){5-7}

        \textbf{Model}
        & \textbf{History}
        & \textbf{No History}
        & \textbf{Partial+}
        & \textbf{History}
        & \textbf{No History}
        & \textbf{Partial+} \\
        \midrule

        \multicolumn{7}{@{}l}{\textit{Vision-Language Models (VLMs)}} \\
        \addlinespace[2pt]

        Qwen3-Omni-30B-A3B
        & 0.62 & \textbf{0.84} & \textbf{19.5\%}
        & 0.89 & \textbf{1.26} & \textbf{37.0\%} \\
        Qwen2.5-VL-7B
        & 0.41 & 0.82 & 17.7\%
        & 0.67 & 1.18 & 31.7\% \\
        Gemma-4-26B-A4B-it
        & \textbf{1.02} & 0.71 & 17.3\%
        & \textbf{1.37} & 0.96 & 24.8\% \\
        Qwen3-VL-8B
        & 0.64 & 0.70 & 15.5\%
        & 0.73 & 1.07 & 28.3\% \\
        Qwen2.5-Omni-7B
        & 0.27 & 0.67 & 13.2\%
        & 0.53 & 0.95 & 24.5\% \\
        Gemma-4-E4B-it
        & 0.77 & 0.66 & 13.8\%
        & 0.94 & 0.96 & 25.7\% \\
        InternVL2.5-8B
        & 0.26 & 0.54 & 10.0\%
        & 0.44 & 0.91 & 27.7\% \\
        Gemma-3-12B-it
        & 0.47 & 0.49 & 10.1\%
        & 0.67 & 0.75 & 22.8\% \\

        \midrule
        \multicolumn{7}{@{}l}{\textit{Computer-Use Agents (CUAs)}} \\
        \addlinespace[2pt]

        UI-TARS-1.5-7B
        & 0.65 & 0.71 & 15.1\%
        & 0.81 & 1.10 & 27.6\% \\
        EvoCUA-8B
        & 0.81 & 0.67 & 15.1\%
        & 1.10 & 0.99 & 25.7\% \\
        OpenCUA-7B
        & 0.63 & 0.59 & 12.0\%
        & 0.85 & 0.97 & 29.3\% \\
        GUI-Owl-1.5-8B
        & 0.56 & 0.56 & 12.1\%
        & 0.89 & 0.79 & 21.0\% \\

        \midrule
        \multicolumn{7}{@{}l}{\textit{Vision-Language-Action Models (VLAs)}} \\
        \addlinespace[2pt]

        OS-Atlas-Pro-7B
        & 0.36 & 0.51 & 10.8\%
        & 0.50 & 0.63 & 15.0\% \\
        ShowUI-2B
        & 0.34 & 0.43 & 8.4\%
        & 0.47 & 0.72 & 21.2\% \\

        \bottomrule
    \end{tabular}
\end{table*}
Table~\ref{tab:overall_results} shows that current multimodal agents make meaningful progress on only a minority of games in \playsuite. In the no-history condition, Qwen3-Omni attains the highest mean Progress Score on both corpora, with 0.84 on \emph{itch.io} and 1.26 on \emph{PyWeek}, while reaching at least \emph{Partial} progress in 19.5\% and 37.0\% of runs, respectively. With interaction history, Gemma-4-26B achieves the highest mean score on both corpora, reaching 1.02 on \emph{itch.io} and 1.37 on \emph{PyWeek}. Performance also does not separate cleanly by model family: CUAs remain competitive with similarly sized VLMs, while the evaluated VLAs generally trail the strongest VLM and CUA models.
\begin{table}[t]
    \centering
    \begin{minipage}[c]{0.40\textwidth}
    \captionsetup{justification=raggedright,singlelinecheck=false}
    \caption{
        \textbf{Progress by genre on \emph{itch.io}.}
        Mean Progress Score and Partial+ rate in the no-history condition for genres with at least 60 games. PyWeek and history results are reported in Appendix~\ref{app:genre-results}.
    }
    \label{tab:genre_results}
    \end{minipage}\hfill
    \begin{minipage}[c]{0.52\textwidth}
    \centering
    \footnotesize
    \renewcommand{\arraystretch}{1.05}
    \begin{tabular}{@{}lrrr@{}}
        \toprule
        Genre      & $N$   & Mean  & Partial+ \\
        \midrule
        Simulation & 506   & 0.893 & 27.1\%   \\
        Strategy   & 223   & 0.858 & 25.2\%   \\
        Shooter    & 205   & 0.771 & 20.4\%   \\
        Survival   & 171   & 0.770 & 21.7\%   \\
        Racing     & 72    & 0.705 & 18.9\%   \\
        Action     & 958   & 0.620 & 14.4\%   \\
        Puzzle     & 1,271 & 0.606 & 10.0\%   \\
        Adventure  & 2,184 & 0.544 & 9.5\%    \\
        \bottomrule
    \end{tabular}
    \end{minipage}
\end{table}

\begin{figure}[t]
    \centering
    \includegraphics[width=.85\textwidth]
        {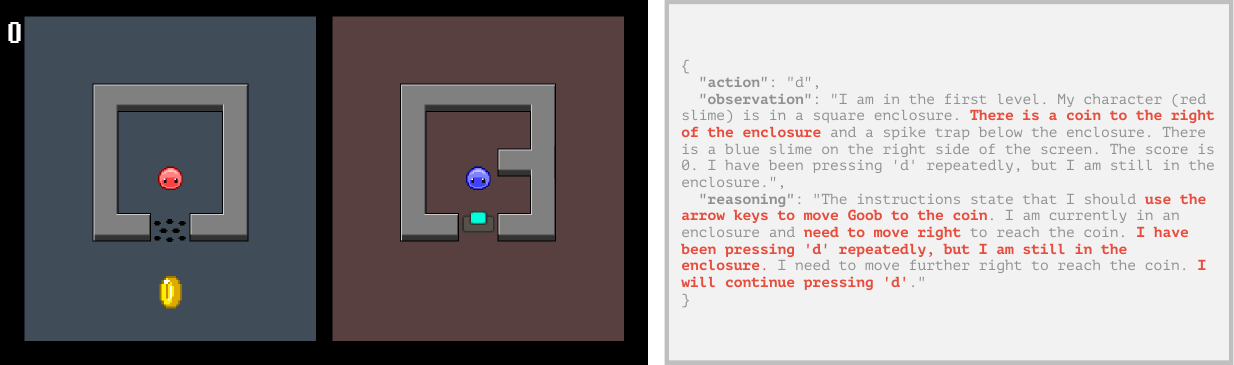}
    \caption{\textbf{Failure mechanisms for Gemma-3 on Goob.}
    The model correctly perceives the position of the spike and blue slime,
    but incorrectly places the coin on the right side of the screen,
    producing a perception error. Although it recognizes that the
    instructions specify the arrow keys for movement, it repeatedly presses
    ``d'' despite receiving no progress, illustrating failures in control
    selection, instruction following, and recovery from repetitive actions.}
    \label{fig:goob-failure}
\end{figure}

\textbf{Qualitative failure modes.}
Inspection of individual trajectories reveals that low progress can arise from failures at multiple stages of the perception-action loop. Common failures include models misreading on-screen text or entities, making inaccurate clicks, or fall into repetitive action loops. Other observed issues include using controls which are unsupported or unmapped by the game, poor instruction following, passive waiting without strategic cause, context pollution from interaction history, and state mismanagement in which previously observed information is contradicted. Figure~\ref{fig:goob-failure} illustrates a few
failure modes within a single trajectory.

\textbf{Performance across game genres.}
    Mean progress varies substantially across game genres. As shown in Table~\ref{tab:genre_results}, among genres containing at least 60 games, models achieve the highest mean progress on Simulation and Strategy games, while Adventure and Puzzle games yield the lowest scores. Despite these differences, the relative ordering of models is largely consistent across genres. The correlation between each genre-specific model ranking and the overall ranking ranges from Spearman $\rho=0.88$ to $0.99$, indicating that models that perform well overall generally also perform well across individual genres. Per-model breakdowns, PyWeek results, and performance with history are provided in Appendix~\ref{app:genre-results}.

\textbf{Effect of action frequency.}
To test whether performance is limited by the 0.33,Hz interaction rate, we repeat the \emph{PyWeek} evaluation at 3,Hz. Mean Progress Score increases from 0.862 to 0.926, indicating a modest benefit from more frequent control. This improvement occurs with and without history, with no clear evidence that action frequency changes the effect of history. See Appendix~\ref{app:rate-results}.

\textbf{How many games are needed?}
To assess how benchmark size affects model comparisons, we repeatedly sample games from the full \emph{itch.io} corpus and compare the resulting no-history model ranking with the full-corpus ranking. With 50 games, the median rank correlation is 0.846, rising to 0.937 with 200 games, 0.969 with 500 games, and 0.987 with 2,000 games. Fine-grained ordering is considerably less stable: the highest-ranked model is recovered in 49.0\% of 50-game samples, 55.5\% at 200 games, 70.0\% at 500 games, and 87.5\% at 2,000 games, while the full-corpus top-three set is recovered in only 26.0\% of 200-game samples and 45.5\% even at 2,000 games. These results show that relatively small samples recover the broad model ordering, whereas substantially larger samples are required to reliably distinguish closely performing models. Accordingly, we use a 200-game \emph{itch.io} subset as a practical scale for the following additional controlled analyses.

\begin{center}
    \centering
    \includegraphics[width=0.98\columnwidth]{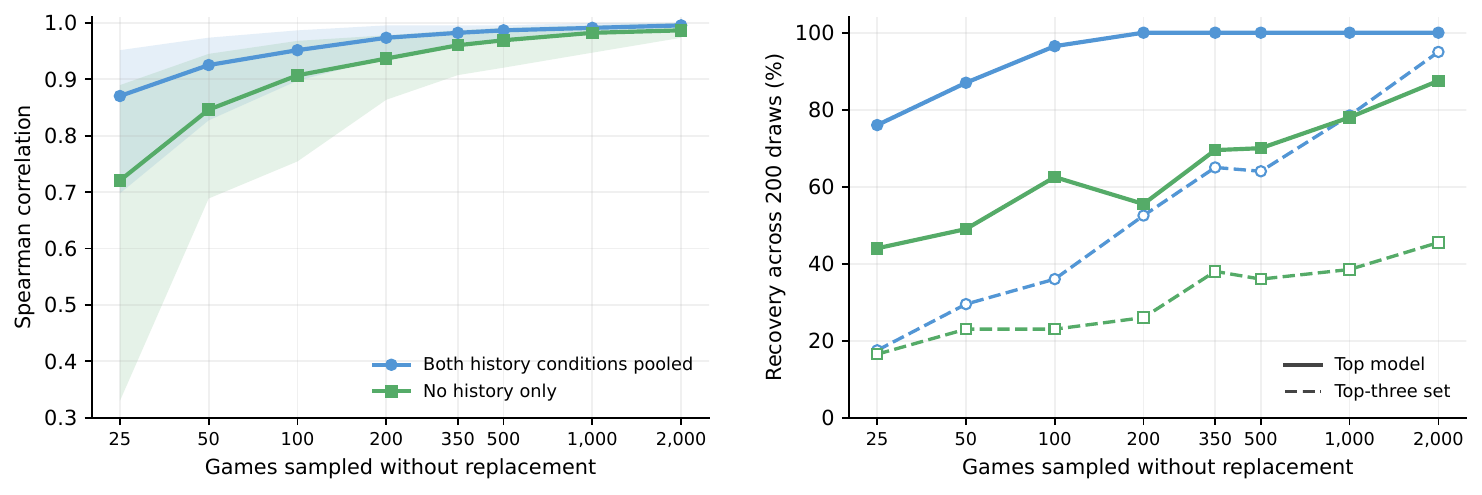}
    \captionof{figure}{\textbf{Leaderboard stability as a function of sampled \emph{itch.io} games.}
    Curves show how model rankings from randomly sampled game subsets compare with the full \emph{itch.io} corpus ranking. Broad model ordering stabilizes with a few hundred games, whereas substantially larger samples are needed for fine-grained comparisons among the highest-ranked models.}
    \label{fig:sample_size}
\end{center}

\textbf{Impact of prompting strategies.}
We test whether explicit strategy traces or few-shot examples improve the eight VLMs evaluated with a shared prompt format, allowing prompt changes to be compared without model-specific interface differences. Neither modification improves average performance, with few-shot prompting reducing progress on both subsets. See Appendix~\ref{app:prompt-results}.

\begin{figure}[t]
    \centering
    \begin{minipage}[c]{0.36\textwidth}
    \centering
    \includegraphics[width=\linewidth]{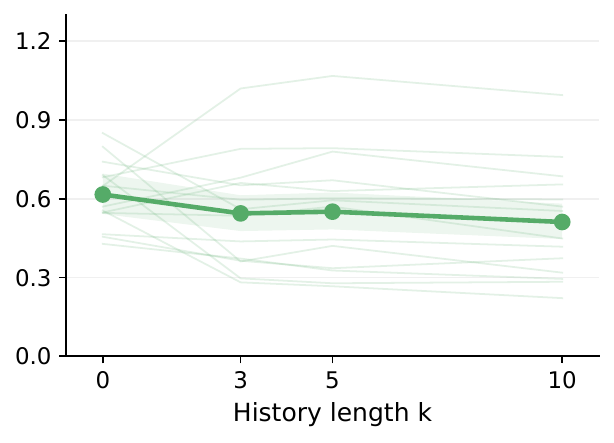}
    \end{minipage}\hspace{0.04\textwidth}
    \begin{minipage}[c]{0.46\textwidth}
    \captionsetup{justification=raggedright,singlelinecheck=false}
    \caption{\textbf{Impact of history length on model performance.} Mean Progress Score across history lengths on the 200-game itch subset. Faint lines show individual models and the bold line shows the mean across models.}
    \label{fig:history_impact}
    \end{minipage}
\end{figure}

\textbf{Impact of temporal history.}
To investigate the role of temporal context, we evaluate model performance across history lengths $k \in \{0,3,5,10\}$ on \emph{PyWeek} and the \emph{itch.io} 200 subset. As illustrated in Figure~\ref{fig:history_impact}, our results reveal a counterintuitive trend: no history yields the highest average progress on both corpora. Increasing history from three to five interactions provides no clear benefit, while extending it to ten interactions reduces progress further. Individual models can nevertheless benefit from history, as shown in Table~\ref{tab:overall_results}. Full results for both subsets are provided in Appendix~\ref{app:history}.

\textbf{Cost-performance analysis}
Evaluation costs vary substantially across models due to model size, serving configuration, and generation length. Higher progress can come at sharply increasing cost: Qwen3-Omni improves mean progress over Qwen2.5-VL, but requires 105.5 versus 59.9 H100 GPU-hours per 1,000 games. Figure~\ref{fig:cost-performance} in Appendix~\ref{app:cost-results} illustrates this cost--performance trade-off.

\section{Conclusion}

We introduce \playsuite, a large-scale benchmark for evaluating interactive visual intelligence in open-source video games. By combining a heterogeneous game corpus, a unified closed-loop interaction harness, scalable cluster execution, and a Video-LLM-as-a-judge evaluation protocol, \playsuite enables systematic measurement of model behavior in settings where agents must translate perception and reasoning into interaction. Strikingly, capabilities that appear strong in static visual reasoning and computer-use settings do not reliably transfer to gameplay: across fourteen open VLMs, CUAs, and VLAs, meaningful progress remains limited to a minority of games, with recurring failures in spatial grounding, control selection, action execution, passive waiting, and the effective use of temporal context.

Recent frontier systems are beginning to demonstrate substantial competence in individual games, making it increasingly important to evaluate whether such capabilities generalize across broad distributions of interactive environments. \playsuite provides a framework for studying this question at scale: our pipeline currently validates approximately 17.5K executable, safe, and reproducible game environments, of which 5,734 are used in the present evaluation. In this sense, \playsuite revisits the broad-environment vision of OpenAI Universe~\citep{openai2016universe} at a time when evaluating generalization across thousands of interactive environments is becoming practically meaningful. As models become increasingly capable of interacting with and generating functional interactive environments, scalable evaluation over large and evolving environment distributions will become increasingly important.

\textbf{Limitations.}
Our evaluation focuses on zero-shot inference and therefore measures out-of-the-box generalization rather than maximum performance after fine-tuning or RL. The closed-loop protocol removes model-inference latency from game dynamics, so \playsuite does not measure real-time reaction speed. Our milestone-based evaluation captures observable progress at scale but does not directly measure strategic efficiency, reasoning quality, or mastery of underlying game mechanics. Finally, because the benchmark uses publicly available games, source code, and metadata, data contamination cannot be ruled out; incorporating newly released games can reduce, but not eliminate, this risk.

\section*{Acknowledgment}
This work was supported by European Union’s Horizon Europe research and innovation programme under grant agreement number 101214398 (ELLIOT). Cees Snoek and Dheeraj Varghese acknowledge EuroHPC Joint Undertaking for awarding the project ID EHPC-AIF-2025SC03-129 access to MareNostrum5 at BSC, Spain. Joaquin Vanschoren and Anna Vettoruzzo acknowledge support from project 2026.008 of the Computing Time National Computing Facilities program which is financed by the Dutch Research Council (NWO) under the grant https://doi.org/10.61686/WBFDE48237. Additionally, Michelle Lorena Acevedo Callejas and Kristof Meding acknowledge funding from the Carl Zeiss Foundation through the Carl Zeiss Institute for AI and Law.

\bibliographystyle{iclr2027_conference}
\bibliography{bibliography}

@inproceedings{he2024webvoyager,
  title={Webvoyager: Building an end-to-end web agent with large multimodal models},
  author={He, Hongliang and Yao, Wenlin and Ma, Kaixin and Yu, Wenhao and Dai, Yong and Zhang, Hongming and Lan, Zhenzhong and Yu, Dong},
  booktitle={Proceedings of the Annual Meeting of the Association for Computational Linguistics},
  pages={6864--6890},
  year={2024}
}

@inproceedings{zhang2025appagent,
  title={Appagent: Multimodal agents as smartphone users},
  author={Zhang, Chi and Yang, Zhao and Liu, Jiaxuan and Li, Yanda and Han, Yucheng and Chen, Xin and Huang, Zebiao and Fu, Bin and Yu, Gang},
  booktitle={Proceedings of the CHI Conference on Human Factors in Computing Systems},
  pages={1--20},
  year={2025}
}

@inproceedings{yue2024mmmu,
  title={Mmmu: A massive multi-discipline multimodal understanding and reasoning benchmark for expert agi},
  author={Yue, Xiang and Ni, Yuansheng and Zhang, Kai and Zheng, Tianyu and Liu, Ruoqi and Zhang, Ge and Stevens, Samuel and Jiang, Dongfu and Ren, Weiming and Sun, Yuxuan and others},
  booktitle={Proceedings of the IEEE/CVF conference on computer vision and pattern recognition},
  pages={9556--9567},
  year={2024}
}

@inproceedings{liu2024mmbench,
  title={Mmbench: Is your multi-modal model an all-around player?},
  author={Liu, Yuan and Duan, Haodong and Zhang, Yuanhan and Li, Bo and Zhang, Songyang and Zhao, Wangbo and Yuan, Yike and Wang, Jiaqi and He, Conghui and Liu, Ziwei and others},
  booktitle={European conference on computer vision},
  pages={216--233},
  year={2024},
  organization={Springer}
}

@article{ying2026ai,
  title={Ai gamestore: Scalable, open-ended evaluation of machine general intelligence with human games},
  author={Ying, Lance and Truong, Ryan and Sharma, Prafull and Zhao, Kaiya Ivy and Cloos, Nathan and Allen, Kelsey R and Griffiths, Thomas L and Collins, Katherine M and Hern{\'a}ndez-Orallo, Jos{\'e} and Isola, Phillip and others},
  journal={arXiv preprint arXiv:2602.17594},
  year={2026}
}

@article{mokaria2026survey,
  title={A Survey on Foundations and Frontiers of Multimodal Agentic Frameworks: Techniques and Applications},
  author={Mokaria, Neel and Raj, Rishie and Baiju, Dheeraj and Shen, Xiaoqian and Pramanick, Shraman and Lin, Kevin Qinghong and Senocak, Arda and Shou, Mike Zheng and Torr, Philip and Elhoseiny, Mohamed and others},
  journal={arXiv preprint arXiv:2608.20379},
  year={2026}
}

@inproceedings{zhouwebarena,
  title={{WebArena}: A Realistic Web Environment for Building Autonomous Agents},
  author={Shuyan Zhou and
Frank F. Xu and
Hao Zhu and
Xuhui Zhou and
Robert Lo and
Abishek Sridhar and
Xianyi Cheng and
Tianyue Ou and
Yonatan Bisk and
Daniel Fried and
Uri Alon and
Graham Neubig},
  booktitle={International Conference on Learning Representations},
  year={2024}
}

@article{lake2017building,
  title={Building machines that learn and think like people},
  author={Lake, Brenden M and Ullman, Tomer D and Tenenbaum, Joshua B and Gershman, Samuel J},
  journal={Behavioral and Brain Sciences},
  volume={40},
  pages={e253},
  year={2017},
  publisher={Cambridge University Press}
}

@inproceedings{lee2014facet,
  title={Facet Analysis of Video Game Genres},
  author={Jin Ha Lee and Natascha Karlova and Rachel Ivy Clarke and Katherine Thornton and Andrew Perti},
  booktitle={Proceedings of the iConference 2014},
  year={2014},
  pages={125-139},
  url={https://api.semanticscholar.org/CorpusID:46322261}
}

@inproceedings{yao2022webshop,
  title={Webshop: Towards scalable real-world web interaction with grounded language agents},
  author={Yao, Shunyu and Chen, Howard and Yang, John and Narasimhan, Karthik},
booktitle={Advances in Neural Information Processing Systems},  volume={35},
  pages={20744--20757},
  year={2022}
}

@inproceedings{yang2025embodiedbench,
  title={{EmbodiedBench}: Comprehensive Benchmarking Multi-modal Large Language Models for Vision-Driven Embodied Agents},
  author={Yang, Rui and Chen, Hanyang and Zhang, Junyu and Zhao, Mark and Qian, Cheng and Wang, Kangrui and Wang, Qineng and Koripella, Teja Venkat and Movahedi, Marziyeh and Li, Manling and others},
  booktitle={International Conference on Machine Learning},
  pages={70576--70631},
  year={2025},
}

@inproceedings{nasir2024gametraversalbenchmark,
  title={{GameTraversalBenchmark}: Evaluating planning abilities of large language models through traversing 2d game maps},
  author={Nasir, Muhammad Umair and James, Steven and Togelius, Julian},
  booktitle={Advances in Neural Information Processing Systems},
  volume={37},
  pages={31813--31827},
  year={2024}
}

@inproceedings{hafner2021benchmarking,
  title={Benchmarking the spectrum of agent capabilities},
  author={Hafner, Danijar},
  booktitle={International Conference on Learning Representations},
  year={2022}
}

@inproceedings{kuttler2020nethack,
  title={The nethack learning environment},
  author={K{\"u}ttler, Heinrich and Nardelli, Nantas and Miller, Alexander and Raileanu, Roberta and Selvatici, Marco and Grefenstette, Edward and Rockt{\"a}schel, Tim},
  booktitle={Advances in Neural Information Processing Systems},
  volume={33},
  pages={7671--7684},
  year={2020}
}

@inproceedings{wang2025large,
  title={Are Large Vision Language Models Good Game Players?},
  author={Wang, Xinyu and Zhuang, Bohan and Wu, Qi},
  booktitle={International Conference on Learning Representations},
  year={2025}
}

@article{zhang2024ing,
  title={{ING-VP}: MLLMs cannot play easy vision-based games yet},
  author={Zhang, Haoran and Guo, Hangyu and Guo, Shuyue and Cao, Meng and Huang, Wenhao and Liu, Jiaheng and Zhang, Ge},
  journal={arXiv preprint arXiv:2410.06555},
  year={2024}
}

@article{qiao2023gameeval,
  title={{GameEval}: Evaluating {LLMs} on conversational games},
  author={Qiao, Dan and Wu, Chenfei and Liang, Yaobo and Li, Juntao and Duan, Nan},
  journal={arXiv preprint arXiv:2308.10032},
  year={2023}
}

@inproceedings{hugamearena,
  title={{GameArena}: Evaluating {LLM} Reasoning through Live Computer Games},
  author={Hu, Lanxiang and Li, Qiyu and Xie, Anze and Jiang, Nan and Stoica, Ion and Jin, Haojian and Zhang, Hao},
  booktitle={International Conference on Learning Representations},
  year={2025}
}

@inproceedings{Xie_Zhang_Chen_Wan_Li_2024, 
title={{WhodunitBench}: Evaluating Large Multimodal Agents via Murder Mystery Games}, 
volume={37}, booktitle={Advances in Neural Information Processing Systems}, author={Xie, Junlin and Zhang, Ruifei and Chen, Zhihong and Wan, Xiang and Li, Guanbin}, year={2024}, pages={86655–86687} }

@inproceedings{fan2022minedojo,
  title={{MineDojo}: Building open-ended embodied agents with internet-scale knowledge},
  author={Fan, Linxi and Wang, Guanzhi and Jiang, Yunfan and Mandlekar, Ajay and Yang, Yuncong and Zhu, Haoyi and Tang, Andrew and Huang, De-An and Zhu, Yuke and Anandkumar, Anima},
  booktitle={Advances in Neural Information Processing Systems},
  volume={35},
  pages={18343--18362},
  year={2022}
}

@inproceedings{Guss_Houghton_Topin_Wang_Codel_Veloso_Salakhutdinov_2019,
  title={{MineRL}: a large-scale dataset of minecraft demonstrations},
  author={Guss, William H and Houghton, Brandon and Topin, Nicholay and Wang, Phillip and Codel, Cayden and Veloso, Manuela and Salakhutdinov, Ruslan},
  booktitle={Proceedings of the International Joint Conference on Artificial Intelligence},
  pages={2442--2448},
  year={2019}
}

@article{zheng2025v,
  title={{V-MAGE}: A Game Evaluation Framework for Assessing Vision-Centric Capabilities in Multimodal Large Language Models},
  author={Zheng, Xiangxi and Li, Linjie and Yang, Zhengyuan and Yu, Ping and Wang, Alex Jinpeng and Yan, Rui and Yao, Yuan and Wang, Lijuan},
  journal={arXiv preprint arXiv:2504.06148},
  year={2025}
}

@inproceedings{paglieribalrog,
  title={{BALROG}: Benchmarking Agentic {LLM} and {VLM} Reasoning On Games},
  author={Paglieri, Davide and Cupia{\l}, Bart{\l}omiej and Coward, Samuel and Piterbarg, Ulyana and Wolczyk, Maciej and Khan, Akbir and Pignatelli, Eduardo and Kuci{\'n}ski, {\L}ukasz and Pinto, Lerrel and Fergus, Rob and others},
  booktitle={International Conference on Learning Representations},
  year={2025}
}

@inproceedings{park2025orak,
  title={Orak: A foundational benchmark for training and evaluating {LLM} agents on diverse video games},
  author={Park, Dongmin and Kim, Minkyu and Choi, Beongjun and Kim, Junhyuck and Lee, Keon and Lee, Jonghyun and Park, Inkyu and Lee, Byeong-Uk and Hwang, Jaeyoung and Ahn, Jaewoo and others},
  booktitle={International Conference on Learning Representations},
  year={2026}
}

@article{wang2026visgym,
  title={{VisGym}: Diverse, Customizable, Scalable Environments for Multimodal Agents},
  author={Wang, Zirui and Zhang, Junyi and Ge, Jiaxin and Lian, Long and Fu, Letian and Dunlap, Lisa and Goldberg, Ken and Wang, XuDong and Stoica, Ion and Chan, David M and others},
  journal={arXiv preprint arXiv:2601.16973},
  year={2026}
}

@article{zhang2025general,
  title={General Modular Harness for {LLM} Agents in Multi-Turn Gaming Environments},
  author={Zhang, Yuxuan and Yu, Haoyang and Hu, Lanxiang and Jin, Haojian and Zhang, Hao},
  journal={arXiv preprint arXiv:2507.11633},
  year={2025}
}

@article{zhang2025videogamebench,
  title={{VideoGameBench}: Can vision-language models complete popular video games?},
  author={Zhang, Alex L and Griffiths, Thomas L and Narasimhan, Karthik R and Press, Ofir},
  journal={arXiv preprint arXiv:2505.18134},
  year={2025}
}

@inproceedings{lu2025cultivating,
  title={Cultivating Game Sense for Yourself: Making {VLMs} Gaming Experts},
  author={Lu, Wenxuan and He, Jiangyang and Zhang, Zhanqiu and Guo, Yiwen and Zang, Tianning},
  booktitle = "Proceedings of the Annual Meeting of the Association for Computational Linguistics",
  year={2025},
  
}

@inproceedings{zheng2023judging,
  title={Judging llm-as-a-judge with mt-bench and chatbot arena},
  author={Zheng, Lianmin and Chiang, Wei-Lin and Sheng, Ying and Zhuang, Siyuan and Wu, Zhanghao and Zhuang, Yonghao and Lin, Zi and Li, Zhuohan and Li, Dacheng and Xing, Eric and others},
  booktitle={Advances in Neural Information Processing Systems},
  volume={36},
  pages={46595--46623},
  year={2023}
}

@inproceedings{pan2024autonomous,
  title={Autonomous Evaluation and Refinement of Digital Agents},
  author={Pan, Jiayi and Zhang, Yichi and Tomlin, Nicholas and Zhou, Yifei and Levine, Sergey and Suhr, Alane},
  booktitle={First Conference on Language Modeling},
  year={2024}
}

@inproceedings{lin2023mcu,
  title={MCU: A Task-centric Framework for Open-ended Agent Evaluation in Minecraft},
  author={Lin, Haowei and Wang, Zihao and Ma, Jianzhu and Liang, Yitao},
  booktitle={Second Agent Learning in Open-Endedness Workshop},
  year={2023}
}

@article{wong2025reasoning,
  title={Reasoning Capabilities of Large Language Models on Dynamic Tasks},
  author={Wong, Annie and B{\"a}ck, Thomas and Plaat, Aske and van Stein, Niki and Kononova, Anna V},
  journal={arXiv preprint arXiv:2505.10543},
  year={2025}
}

@inproceedings{xu2025vs,
  title={{VS-Bench}: Evaluating {VLMs} for Strategic Abilities in Multi-Agent Environments},
  author={Xu, Zelai and Xu, Zhexuan and Yi, Xiangmin and Yuan, Huining and Guang, Mo and Long, Kaiwen and Chen, Xinlei and Wu, Yi and Yu, Chao and Wang, Yu},
  booktitle = {Proceedings of the IEEE Conference on Computer Vision and Pattern Recognition},
  year={2026}
}

@misc{pyweek,
  author = {{PyWeek}},
  title = {Python Game Programming Challenge},
  howpublished = {\url{https://pyweek.org}},
  note = {Accessed: 2026-04-28}
}

@misc{itchio,
  author = {{Itch.io}},
  title = {A simple way to find, download and distribute indie games online},
  howpublished = {\url{https://itch.io}},
  note = {Accessed: 2026-04-28}
}

@inproceedings{chen2025recent,
  title={Recent advances in large language model benchmarks against data contamination: From static to dynamic evaluation},
  author={Chen, Simin and Chen, Yiming and Li, Zexin and Jiang, Yifan and Wan, Zhongwei and He, Yixin and Ran, Dezhi and Gu, Tianle and Li, Haizhou and Xie, Tao and others},
    booktitle = "Proceedings of the Conference on Empirical Methods in Natural Language Processing",
    year = "2025"
}

@article{bai2025qwen2,
  title={{Qwen2.5-VL} Technical Report},
  author={Bai, Shuai and Chen, Keqin and Liu, Xuejing and Wang, Jialin and Ge, Wenbin and Song, Sibo and Dang, Kai and Wang, Peng and Wang, Shijie and Tang, Jun and others},
  journal={arXiv preprint arXiv:2502.13923},
  year={2025}
}

@article{bai2025qwen3,
  title={{Qwen3-VL} technical report},
  author={Bai, Shuai and Cai, Yuxuan and Chen, Ruizhe and Chen, Keqin and Chen, Xionghui and Cheng, Zesen and Deng, Lianghao and Ding, Wei and Gao, Chang and Ge, Chunjiang and others},
  journal={arXiv preprint arXiv:2511.21631},
  year={2025}
}

@article{gemma2025, 
title={Gemma 3 Technical Report}, author={{Gemma Team}}, journal={arXiv preprint arXiv:2503.19786}, year={2025} }

@misc{gemmateam2026gemma4,
      title={Gemma 4 Technical Report}, 
      author={{Gemma Team}},
      year={2026},
      eprint={2607.02770},
      archivePrefix={arXiv},
      primaryClass={cs.CL},
      url={https://arxiv.org/abs/2607.02770}
}

@misc{xu2025qwen3omnitechnicalreport,
  title={Qwen3-Omni Technical Report}, 
  author={Jin Xu and Zhifang Guo and Hangrui Hu and Yunfei Chu and Xiong Wang and Jinzheng He and Yuxuan Wang and Xian Shi and Ting He and Xinfa Zhu and Yuanjun Lv and Yongqi Wang and Dake Guo and He Wang and Linhan Ma and Pei Zhang and Xinyu Zhang and Hongkun Hao and Zishan Guo and Baosong Yang and Bin Zhang and Ziyang Ma and Xipin Wei and Shuai Bai and Keqin Chen and Xuejing Liu and Peng Wang and Mingkun Yang and Dayiheng Liu and Xingzhang Ren and Bo Zheng and Rui Men and Fan Zhou and Bowen Yu and Jianxin Yang and Le Yu and Jingren Zhou and Junyang Lin},
  year={2025},
  eprint={2509.17765},
  archivePrefix={arXiv},
  primaryClass={cs.CL},
  url={https://arxiv.org/abs/2509.17765}
}

@misc{xu2025qwen25omnitechnicalreport,
  title={Qwen2.5-Omni Technical Report}, 
  author={Jin Xu and Zhifang Guo and Jinzheng He and Hangrui Hu and Ting He and Shuai Bai and Keqin Chen and Jialin Wang and Yang Fan and Kai Dang and Bin Zhang and Xiong Wang and Yunfei Chu and Junyang Lin},
  year={2025},
  eprint={2503.20215},
  archivePrefix={arXiv},
  primaryClass={cs.CL},
  url={https://arxiv.org/abs/2503.20215} 
}

@article{internvl,
  title={Expanding performance boundaries of open-source multimodal models with model, data, and test-time scaling},
  author={Chen, Zhe and Wang, Weiyun and Cao, Yue and Liu, Yangzhou and Gao, Zhangwei and Cui, Erfei and Zhu, Jinguo and Ye, Shenglong and Tian, Hao and Liu, Zhaoyang and others},
  journal={arXiv preprint arXiv:2412.05271},
  year={2024}
}

@misc{evocua,
  title={EvoCUA: Evolving Computer Use Agents via Learning from Scalable Synthetic Experience}, 
  author={Taofeng Xue and Chong Peng and Mianqiu Huang and Linsen Guo and Tiancheng Han and Haozhe Wang and Jianing Wang and Xiaocheng Zhang and Xin Yang and Dengchang Zhao and Jinrui Ding and Xiandi Ma and Yuchen Xie and Peng Pei and Xunliang Cai and Xipeng Qiu},
  year={2026},
  eprint={2601.15876},
  archivePrefix={arXiv},
  primaryClass={cs.AI},
  url={https://arxiv.org/abs/2601.15876}
}

@inproceedings{opencua,
 author = {Wang, Xinyuan and Wang, Bowen and Lu, Dunjie and Yang, Junlin and Xie, Tianbao and Wang, Junli and Deng, Jiaqi and Guo, Xiaole and Xu, Yiheng and Wu, Chen and Shen, Zhennan and Li, Zhuokai and Li, Ryan and Li, Xiaochuan and Chen, Junda and Zheng, Boyuan and PEIHANG, LI and Lei, Fangyu and Cao, Ruisheng and Fu, Yeqiao and Shin, Dongchan and Shin, Martin and Jiarui, Hu and Wang, Yuyan and Chen, Jixuan and Ye, Yuxiao and Zhang, Danyang and Wang, Yipu and Wang, Heng and Yang, Diyi and Zhong, Victor and Y.Charles and Yang, Zhilin and Yu, Tao},
 booktitle = {Advances in Neural Information Processing Systems},
 doi = {10.52202/085713-4669},
 editor = {D. Belgrave and C. Zhang and H. Lin and R. Pascanu and P. Koniusz and M. Ghassemi and N. Chen},
 pages = {139756--139806},
 publisher = {Curran Associates, Inc.},
 title = {{OpenCUA}: Open Foundations for Computer-Use Agents},
 url = {https://proceedings.neurips.cc/paper_files/paper/2025/file/cc7ae529e945226b0d52ea4ac478c4f3-Paper-Conference.pdf},
 volume = {38},
 year = {2025}
}

@misc{gui-owl-15,
  title={Mobile-Agent-v3.5: Multi-platform Fundamental GUI Agents}, 
  author={Haiyang Xu and Xi Zhang and Haowei Liu and Junyang Wang and Zhaozai Zhu and Shengjie Zhou and Xuhao Hu and Feiyu Gao and Junjie Cao and Zihua Wang and Zhiyuan Chen and Jitong Liao and Qi Zheng and Jiahui Zeng and Ze Xu and Shuai Bai and Junyang Lin and Jingren Zhou and Ming Yan},
  year={2026},
  eprint={2602.16855},
  archivePrefix={arXiv},
  primaryClass={cs.AI},
  url={https://arxiv.org/abs/2602.16855}
}

@article{qin2025ui,
  title={{UI-TARS}: Pioneering automated gui interaction with native agents},
  author={Qin, Yujia and Ye, Yining and Fang, Junjie and Wang, Haoming and Liang, Shihao and Tian, Shizuo and Zhang, Junda and Li, Jiahao and Li, Yunxin and Huang, Shijue and others},
  journal={arXiv preprint arXiv:2501.12326},
  year={2025}
}

@inproceedings{lin2025showui,
  title={Showui: One vision-language-action model for gui visual agent},
  author={Lin, Kevin Qinghong and Li, Linjie and Gao, Difei and Yang, Zhengyuan and Wu, Shiwei and Bai, Zechen and Lei, Stan Weixian and Wang, Lijuan and Shou, Mike Zheng},
  booktitle={Proceedings of the Computer Vision and Pattern Recognition Conference},
  pages={19498--19508},
  year={2025}
}

@inproceedings{wuatlas,
  title={{OS-ATLAS}: Foundation Action Model for Generalist GUI Agents},
  author={Wu, Zhiyong and Wu, Zhenyu and Xu, Fangzhi and Wang, Yian and Sun, Qiushi and Jia, Chengyou and Cheng, Kanzhi and Ding, Zichen and Chen, Liheng and Liang, Paul Pu and others},
  booktitle={International Conference on Learning Representations},
  year={2025}
}

@inproceedings{yao2023react,
  title={{ReAct}: Synergizing Reasoning and Acting in Language Models},
  author={Yao, Shunyu and Zhao, Jeffrey and Yu, Dian and Du, Nan and Shafran, Izhak and Narasimhan, Karthik and Cao, Yuan},
  booktitle={International Conference on Learning Representations},
  year={2023}
}

@inproceedings{huang2022language,
  title={Language models as zero-shot planners: Extracting actionable knowledge for embodied agents},
  author={Huang, Wenlong and Abbeel, Pieter and Pathak, Deepak and Mordatch, Igor},
  booktitle={International Conference on Machine Learning},
  pages={9118--9147},
  year={2022}
}

@inproceedings{costarelli2024gamebench,
  title={GameBench: Evaluating Strategic Reasoning Abilities of {LLM} Agents},
  author={Costarelli, Anthony and Allen, Mat and Hauksson, Roman and Sodunke, Grace and Hariharan, Suhas and Cheng, Carlson and Li, Wenjie and Clymer, Joshua M and Yadav, Arjun},
  booktitle={Language Gamification Workshop},
  year={2024}
}

@article{xie2023translating,
  title={Translating natural language to planning goals with large-language models},
  author={Xie, Yaqi and Yu, Chen and Zhu, Tongyao and Bai, Jinbin and Gong, Ze and Soh, Harold},
  journal={arXiv preprint arXiv:2302.05128},
  year={2023}
}

@article{wang2024survey,
  title={A survey on large language model based autonomous agents},
  author={Wang, Lei and Ma, Chen and Feng, Xueyang and Zhang, Zeyu and Yang, Hao and Zhang, Jingsen and Chen, Zhiyuan and Tang, Jiakai and Chen, Xu and Lin, Yankai and others},
  journal={Frontiers of Computer Science},
  volume={18},
  number={6},
  pages={186345},
  year={2024},
  publisher={Springer}
}

@inproceedings{wang2023describe,
  title={Describe, explain, plan and select: interactive planning with large language models enables open-world multi-task agents},
  author={Wang, Zihao and Cai, Shaofei and Chen, Guanzhou and Liu, Anji and Ma, Xiaojian and Liang, Yitao},
  booktitle={Advances in Neural Information Processing Systems},
  pages={34153--34189},
  year={2023}
}

@article{chollet2025arc,
  title={{ARC AGI 2}: A new challenge for frontier ai reasoning systems},
  author={Chollet, Francois and Knoop, Mike and Kamradt, Gregory and Landers, Bryan and Pinkard, Henry},
  journal={arXiv preprint arXiv:2505.11831},
  year={2025}
}

@article{foundation2026arc,
  title={{ARC-AGI-3}: A New Challenge for Frontier Agentic Intelligence},
  author={{ARC Prize Foundation}},
  journal={arXiv preprint arXiv:2603.24621},
  year={2026}
}

@article{hu2024survey,
  title={A survey on large language model-based game agents},
  author={Hu, Sihao and Huang, Tiansheng and Liu, Gaowen and Kompella, Ramana Rao and Ilhan, Fatih and Tekin, Selim Furkan and Xu, Yichang and Yahn, Zachary and Liu, Ling},
  journal={arXiv preprint arXiv:2404.02039},
  year={2024}
}

@misc{steamdb,
  author = {{SteamDB}},
  title = {Database of everything on Steam},
  howpublished = {\url{https://steamdb.info/}},
  note = {Accessed: 2026-04-28}
}

@article{xie2024osworld,
  title={Osworld: Benchmarking multimodal agents for open-ended tasks in real computer environments},
  author={Xie, Tianbao and Zhang, Danyang and Chen, Jixuan and Li, Xiaochuan and Zhao, Siheng and Cao, Ruisheng and Hua, Toh J and Cheng, Zhoujun and Shin, Dongchan and Lei, Fangyu and others},
  journal={Advances in Neural Information Processing Systems},
  volume={37},
  pages={52040--52094},
  year={2024}
}

@misc{anthropic2026fable,
  author       = {{Anthropic}},
  title        = {Claude Fable 5 and Claude Mythos 5},
  year         = {2026},
  month        = jun,
  howpublished = {Anthropic},
  note         = {June 9, 2026}
}

@article{bolton2025sima,
  title={Sima 2: A generalist embodied agent for virtual worlds},
  author={Bolton, Adrian and Lerchner, Alexander and Cordell, Alexandra and Moufarek, Alexandre and Bolt, Andrew and Lampinen, Andrew and Mitenkova, Anna and Hallingstad, Arne Olav and Vujatovic, Bojan and Li, Bonnie and others},
  journal={arXiv preprint arXiv:2512.04797},
  year={2025}
}

@misc{openai2026astra,
  author       = {{OpenAI}},
  title        = {{GPT-6 Astra}: A New Generation of Intelligence},
  year         = {2026},
  howpublished = {OpenAI}
}

@misc{intelligence2025pi_,
      title={$\pi_{0.5}$: a Vision-Language-Action Model with Open-World Generalization}, 
      author={{Physical Intelligence} and Kevin Black and Noah Brown and James Darpinian and Karan Dhabalia and Danny Driess and Adnan Esmail and Michael Equi and Chelsea Finn and Niccolo Fusai and Manuel Y. Galliker and Dibya Ghosh and Lachy Groom and Karol Hausman and Brian Ichter and Szymon Jakubczak and Tim Jones and Liyiming Ke and Devin LeBlanc and Sergey Levine and Adrian Li-Bell and Mohith Mothukuri and Suraj Nair and Karl Pertsch and Allen Z. Ren and Lucy Xiaoyang Shi and Laura Smith and Jost Tobias Springenberg and Kyle Stachowicz and James Tanner and Quan Vuong and Homer Walke and Anna Walling and Haohuan Wang and Lili Yu and Ury Zhilinsky},
      year={2025},
      eprint={2504.16054},
      archivePrefix={arXiv},
      primaryClass={cs.LG},
      url={https://arxiv.org/abs/2504.16054}, 
}

@article{zhou2026qwen,
  title={Qwen-UI-Agent Technical Report: Toward Next-Generation Real-World Centric Foundation GUI Agents},
  author={Zhou, Hanzhang and Tong, Panrong and Zhang, Xu and Kong, Quyu and Cai, Chenglin and Xia, Tianyu and Zhang, Gongjie and Zhang, Jianan and Li, Long and Chen, Long and others},
  journal={arXiv preprint arXiv:2607.28227},
  year={2026}
}

@article{tan2024cradle,
  title={Cradle: Empowering foundation agents towards general computer control},
  author={Tan, Weihao and Zhang, Wentao and Xu, Xinrun and Xia, Haochong and Ding, Ziluo and Li, Boyu and Zhou, Bohan and Yue, Junpeng and Jiang, Jiechuan and Li, Yewen and others},
  journal={arXiv preprint arXiv:2403.03186},
  year={2024}
}

@misc{openai2016universe,
  author       = {{OpenAI}},
  title        = {Universe},
  year         = {2016},
  month        = dec,
  day          = {5},
  howpublished = {OpenAI},
  url          = {https://openai.com/index/universe/}
}

\clearpage
\appendix
\section{Appendix}

\subsection{Data curation methodology and technical validation}
\label{app:data_curation}
The transition from the raw data pool to the final \playsuite benchmark involved a rigorous multi-stage pipeline focused on quality, technical viability, and safety. We initially considered a pool of 1460 \emph{PyWeek} games and 1.7M \emph{itch.io} games. Across both platforms, we applied a quality gate by selecting only games with a community rating $> 3.0/5.0$. For the \emph{itch.io} corpus specifically, we applied additional constraints, including only free-to-play and fully-released titles with at least 5 individual ratings, verified compatibility for Linux or Web-based execution, support for keyboard and/or mouse input. We additionally executed each candidate game to confirm it launches without errors. Following this initial filtering, we applied dataset-specific refinement layers. The \emph{PyWeek} dataset underwent three further rounds of technical pruning. We excluded titles requiring deprecated Python versions (e.g., Python 2.x), leaving 376 video games. We subsequently restricted the selection to games developed using the \texttt{Pygame} library and performed automated dry-runs to identify and remove games containing internal runtime bugs or broken assets. This resulted in a final, stable set of 104 high-fidelity Python games.

The \emph{itch.io} dataset underwent a dual-layered security and content audit to mitigate risks associated with executing third-party binaries. First, we implemented an automated content filtering to exclude games containing NSFW content, via tag-based matching. We also conducted an additional targeted analysis with the specific intent of identifying games where an AI agent could plausibly encounter situations matching particular risk categories. This allowed us to categorize potential safety-critical interactions within the benchmark, as reported in Table~\ref{tab:harmful_material_stats}. Candidate games were identified via tag-based matching and validated using~\emph{gpt-5.4-nano-2026-03-17}. Second, we deployed a custom safety triage scanner to evaluate structural game archive signals (e.g., password protection, nested binaries) and metadata risks. Archives were assigned an integer risk score to gate our execution policy.
High-risk games (score $\ge 4$) were quarantined for manual review, while games with a score of 3, often indicating download failures or unexplained binaries, were subject to extra logging. Minor signals (scores 1–2) and clean archives (score 0) were executed within our network-isolated sandbox. Established games with high ratings (rating count > 50) were categorized as ``trusted'' (score $< 0$). Following these filters, the final \textit{itch.io} collection was pruned to 5,630 games, ensuring a secure and computationally reliable distribution for large-scale evaluation. Following technical validation, we excluded games that could not be evaluated reliably. Table~\ref{tab:technical_exclusions} summarizes these exclusion reasons.

\begin{table}[!htbp]
    \centering
    \caption{Distribution of harmful content categories identified in the \emph{itch.io} dataset.}
    \label{tab:harmful_material_stats}
    \begin{tabular}{lr}
        \toprule
        \textbf{Harmful Categories} & \textbf{Game Count} \\
        \midrule
        Firearms Individual Violence & 13 \\
        Scheming Manipulation        & 35 \\
        Value Misalignment           & 91 \\
        Inconsistent Behavior        & 24 \\
        Explosives Violence          & 3  \\
        Economic Waste               & 12 \\
        Data Exfiltration            & 4  \\
        Economic Redistribution      & 2  \\
        Explosives Creation          & 4  \\
        Infrastructure Damage        & 4  \\
        Firearms Mass Violence       & 2  \\
        Bias Stereotypes             & 4  \\
        Accidents                    & 2  \\
        \bottomrule
    \end{tabular}
\end{table}

\begin{table}[!htbp]
\centering
\small
\caption{\textbf{Additional technical exclusions from the evaluation set.}
Number of games excluded and their identified reason.}
\label{tab:technical_exclusions}
\begin{tabular}{@{}lr@{}}
\toprule
\textbf{Reason} & \textbf{Games} \\
\midrule
Construct runtime freeze failure & 225 \\
Dragging is the only listed control & 72 \\
Requires multiple players & 25 \\
Not an interactive game & 18 \\
Gamepad required & 2 \\
\bottomrule
\end{tabular}
\end{table}

\subsection{Game genre taxonomy and categorization}
\label{app:taxonomy_categories}
The taxonomy used in \playsuite was developed through a multi-stage process of data-driven analysis and alignment with established game-AI literature. Our goal was to create a classification system that reflects both the computational challenges (e.g., reactive control vs. long-term reasoning) and the distribution of mechanics present in the source platforms.

We began by analyzing the native metadata from the \emph{itch.io} crawl. After initial filtering, we identified 17 primary category tags: \texttt{Action, Platformer, Adventure, Puzzle, Shooter, Simulation, Game Roly Playing, Survival, Strategy, Visual Novel, Interactive Fiction, Educational, Fighting, Racing, Card Game, Rythm, Sports}. However, we observed that individual titles were frequently mapped to multiple tags (e.g., a \texttt{Platformer} being tagged as both \texttt{Action} and \texttt{Adventure}). To resolve these ambiguities and ensure the utility for evaluating specific AI capabilities, we cross-referenced these tags with established gameplay taxonomies. Specifically, our categorization is grounded in recent literature on LLM-based game agents, which distinguishes genres based on the core cognitive challenges and agent capabilities required, such as reasoning-heavy vs. reaction-based dynamics~\citep{hu2024survey, Lee2014facet}.

By reconciling these academic perspectives with industry-standard tagging systems like \textit{SteamDB}~\citep{steamdb}, we consolidated the library into 9 distinct genres. Crucially, we formulated detailed qualitative descriptions for each category, following~\citep{Lee2014facet}. These descriptions serve a dual purpose: providing a rigorous definition for the taxonomy and acting as the semantic grounding for our automated classification of untagged games.
\begin{itemize}
    \item \texttt{Action}: Games with a heavy emphasis on a series of actions performed by the player in order to meet a certain set of objectives. Examples are Metal Gear Solid, Rhythm Doctor, Super Mario Bros and Space Invaders.
    \item \texttt{Adventure}: Games which are set in a world for the player to explore and complete a certain set of objectives through a series of actions. Examples include The Legend of Zelda and Prince of Persia.
    \item \texttt{Puzzle}: Games with an objective of figuring out the solution by solving enigmas, navigating, and manipulating and reconfiguring objects. Examples include Tetris and Minesweeper.
    \item \texttt{Shooter}: Games involving shooting at, and often destroying, a series of opponents or objects. Examples include Doom and Duck Hunt.
    \item \texttt{Simulation}: Games recreating an experience of a real world activity in the game world. Examples include SimCity, The FIFA series, Trauma Center.
    \item \texttt{Survival}: Survival games are action-based, open-world experiences where players start with minimal resources in a hostile environment, tasked with staying alive as long as possible. Examples include Minecraft and The Forest.
    \item \texttt{Strategy}: Games characterized by players' strategic decisions and interventions to bring the desired outcome. Examples include StarCraft and the Total War series.
    \item \texttt{Fighting}: Games where the player controls a game character to engage in a combat against an opponent. Examples include Street Fighter and Mortal Kombat.
    \item \texttt{Racing}: Games involving driving various types of vehicles as the main action, with an objective of winning a race against an opponent. Examples include Mario Kart and Gran Turismo.
\end{itemize}

Following our taxonomy, we mapped the game library to these genres based on platform-specific metadata.

\begin{table}[t]
\caption{Mapping of original \emph{itch.io} metadata tags to the 9 canonical \playsuite genres.}
\label{tab:genre_mapping}
\centering
\begin{tabular}{ll}
\toprule
\textbf{Original \emph{itch.io} genres} & \textbf{\playsuite genre}\\
\midrule
Puzzle, Educational & Puzzle \\
Adventure, Visual Novel, Interactive Fiction, Role Playing & Adventure \\
Action, Platformer, Rhythm & Action \\
Simulation, Sports & Simulation \\
Strategy, Card Game & Strategy \\
Shooter & Shooter \\
Survival & Survival \\
Racing & Racing \\
Fighting & Fighting \\
\bottomrule
\end{tabular}
\end{table}

For the \textit{itch.io} dataset, we performed a rule-based merge to map the 17 original categories into our 9 canonical genres, as shown in Table~\ref{tab:genre_mapping}. When a game carried multiple tags, we resolved them to a single canonical genre by selecting the more specific \playsuite category, where specificity is measured as the inverse number of \textit{itch.io} tags mapped onto each \playsuite genre (e.g., \texttt{Racing}, mapped from a single \textit{itch.io} tag, is more specific than \texttt{Action}, which is mapped from three). If a game had no genre metadata, we assigned one using~\emph{gpt-5.4-nano-2026-03-17}, providing the model with the game title, tags and description.

\begin{figure}[!htbp]
\centering
\includegraphics[width=\linewidth]{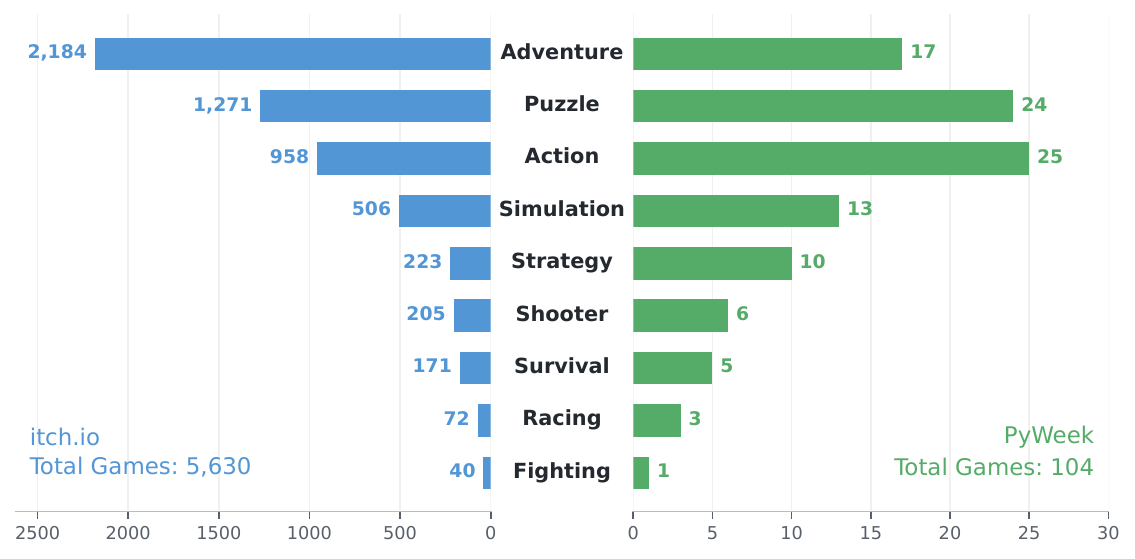}
\caption{\textbf{Genre distribution of the \playsuite evaluation set.} Distribution of the 5,630 evaluated \emph{itch.io} games and 104 \emph{PyWeek} games across the nine canonical \playsuite genres.}
\label{fig:game_categories_comparison}
\end{figure}

For the \textit{PyWeek} dataset, which lacks a native tagging system, we implemented an automated classification pipeline using Qwen2.5-14B-Instruct. To ensure consistency and high-fidelity mapping, we provided the model with the exact category descriptions and representative examples detailed in our taxonomy. For each game, the model analyzed the provided game description, README file and diary entries to infer the primary genre. By using the qualitative definitions of our taxonomy as the direct prompt, we ensured that the \emph{PyWeek} genre categorization remained semantically aligned with the \emph{itch.io} set, creating a unified distribution across the entire benchmark. Specifically, the textual definitions for the 9 categories provided above were used verbatim to populate the \texttt{<categories\_str>} placeholder in our classification prompt. The specific prompt template used for the automated categorization process is detailed in Figure~\ref{fig:prompt_categorization}. The resulting genre distributions for the evaluated \emph{itch.io} and \emph{PyWeek} subsets are shown in Figure~\ref{fig:game_categories_comparison}.

\begin{figure}[htbp]
\begin{mdframed}[backgroundcolor=gray!10, linecolor=gray!50, innerleftmargin=10pt, innerrightmargin=10pt, innertopmargin=10pt, innerbottommargin=10pt]
\small
\textbf{Prompt for PyWeek Categorization:} \\[\baselineskip]
You are an expert in videogame genre classification. Your task is to assign each game to EXACTLY ONE of the genres in the following list:

\texttt{<categories\_str>}

In the list above, each entry corresponds to one genre. For each entry, the name of the genre is before the colon and the genre description is after the colon.
The description also includes examples of games in that genre.
\\[\baselineskip]
Analyze the game information below and determine the most appropriate genre.

Game Name: \texttt{<game\_name>}

Game Description: \texttt{<game\_description>}  \texttt{<readme\_file>}

Developer Diary Entries: \texttt{<diary\_entries>}

Based on the game name, description, and developer notes above, determine the PRIMARY genre.
\\[\baselineskip]
IMPORTANT RULES:

1. Respond with ONLY the genre name from the list above

2. Choose the single BEST matching genre

3. If unsure, make your best guess based on the description

4. Remember that a video game genre is defined by similar gameplay characteristics and how the player interacts with the environment, rather than by the setting or story of the game or its medium of play.

5. In your reasoning, include two to three reasons why your selection genre is the best match for the game. Please indicate what information you used to infer these reasons. If you have low confidence in the category selection, please also add a sentence or two explaining why you are unsure.

6. Return ONLY a JSON object with this format:

\{
  "categories": ["Category"],

  "confidence": "high|medium|low",

  "reasoning": "Brief explanation"
\}
\\[\baselineskip]
Example output:
\{"categories": ["Action"],
"confidence": "high",
"reasoning": "The game involves jumping between platforms and combat mechanics. As mentioned in the read me, the objective is to take the player from starting point A to end point B. Diary entry 2 shows that there are X levels in the game."\}
\\[\baselineskip]
Your JSON response:

\end{mdframed}
\caption{Prompt used for the automated categorization of PyWeek games.} \label{fig:prompt_categorization}
\end{figure}

\subsection{Licensing and Terms of Use}
\label{app:licensing}
\playsuite respects the intellectual property rights and distribution terms set forth by the original hosting platforms and developers.

In accordance with the official \emph{PyWeek} rules~\citep{pyweek}, developers retain all copyrights to their entries. However, by submitting an entry, authors grant a transferable, irrevocable license to redistribute, copy, and run the entry without modification, and to distribute unmodified screenshots, provided no fee is charged. We distribute the \emph{PyWeek} subset of \playsuite under these terms, ensuring that the original game files remain unmodified and that our benchmark is provided as a free resource for the research community. For games that included independent license files within their source code, those specific terms are respected as an alternative to the standard \emph{PyWeek} clause.

Games sourced from \emph{itch.io} were selected from publicly available, free-to-play titles. In accordance with the \emph{itch.io} Terms of Service~\citep{itchio}, developers retain ownership of the content they upload. Our evaluation framework interacts with these games as a ``user'' of the service, utilizing the files as provided by the developers for the purpose of non-commercial research and benchmarking. We do not modify the game binaries or assets. To ensure compliance with developer intent, we have restricted our corpus to titles that are explicitly marked for public distribution and free access. Furthermore, we provide metadata and direct links to the original game pages to ensure proper attribution to the original creators.

\subsection{Model configurations}
\label{app:models}
Table~\ref{tab:baseline_models} summarizes the 14 evaluated baselines, their model families, parameter scales, and checkpoint identifiers.

\begin{table}[!htbp]
\centering
\footnotesize
\caption{\textbf{Baseline models.} Model family, parameter scale, Hugging Face checkpoint, and reference for each
evaluated model.}
\label{tab:baseline_models}
\begingroup
\setlength{\tabcolsep}{3pt}
\def\UrlBreaks{\do\/\do-\do\_\do.}
\renewcommand{\arraystretch}{1.18}
\begin{tabular}{@{}p{.22\linewidth}p{.07\linewidth}p{.125\linewidth}p{.33\linewidth}p{\dimexpr.255\linewidth-24pt\relax}@{}}
\toprule
\textbf{Model} & \textbf{Family} & \textbf{Parameters} & \textbf{Hugging Face identifier} & \textbf{Reference}\\
\midrule
\raggedright Qwen3-Omni-30B-A3B-Instruct & VLM & \raggedright 30B\newline (3B active) & \raggedright\href{https://huggingface.co/Qwen/Qwen3-Omni-30B-A3B-Instruct}{\nolinkurl{Qwen/Qwen3-Omni-30B-A3B-Instruct}} & \citep{xu2025qwen3omnitechnicalreport} \\
\raggedright Qwen2.5-VL-7B-Instruct & VLM & \raggedright 7B & \raggedright\href{https://huggingface.co/Qwen/Qwen2.5-VL-7B-Instruct}{\nolinkurl{Qwen/Qwen2.5-VL-7B-Instruct}} & \citep{bai2025qwen2} \\
\raggedright Gemma-4-26B-A4B-it & VLM & \raggedright 26B\newline (4B active) & \raggedright\href{https://huggingface.co/google/gemma-4-26B-A4B-it}{\nolinkurl{google/gemma-4-26B-A4B-it}} & \citep{gemmateam2026gemma4} \\
\raggedright Qwen3-VL-8B-Instruct & VLM & \raggedright 8B & \raggedright\href{https://huggingface.co/Qwen/Qwen3-VL-8B-Instruct}{\nolinkurl{Qwen/Qwen3-VL-8B-Instruct}} & \citep{bai2025qwen3} \\
\raggedright Qwen2.5-Omni-7B & VLM & \raggedright 7B & \raggedright\href{https://huggingface.co/Qwen/Qwen2.5-Omni-7B}{\nolinkurl{Qwen/Qwen2.5-Omni-7B}} & \citep{xu2025qwen25omnitechnicalreport} \\
\raggedright Gemma-4-E4B-it & VLM & \raggedright 8B\newline (4.5B eff.) & \raggedright\href{https://huggingface.co/google/gemma-4-E4B-it}{\nolinkurl{google/gemma-4-E4B-it}} & \citep{gemmateam2026gemma4} \\
\raggedright InternVL2.5-8B & VLM & \raggedright 8B & \raggedright\href{https://huggingface.co/OpenGVLab/InternVL2_5-8B}{\nolinkurl{OpenGVLab/InternVL2_5-8B}} & \citep{internvl} \\
\raggedright Gemma-3-12B-it & VLM & \raggedright 12B & \raggedright\href{https://huggingface.co/google/gemma-3-12b-it}{\nolinkurl{google/gemma-3-12b-it}} & \citep{gemma2025} \\
\midrule
\raggedright UI-TARS-1.5-7B & CUA & \raggedright 7B & \raggedright\href{https://huggingface.co/ByteDance-Seed/UI-TARS-1.5-7B}{\nolinkurl{ByteDance-Seed/UI-TARS-1.5-7B}} & \citep{qin2025ui} \\
\raggedright EvoCUA-8B & CUA & \raggedright 8B & \raggedright\href{https://huggingface.co/meituan/EvoCUA-8B-20260105}{\nolinkurl{meituan/EvoCUA-8B-20260105}} & \citep{evocua} \\
\raggedright OpenCUA-7B & CUA & \raggedright 7B & \raggedright\href{https://huggingface.co/xlangai/OpenCUA-7B}{\nolinkurl{xlangai/OpenCUA-7B}} & \citep{opencua} \\
\raggedright GUI-Owl-1.5-8B-Instruct & CUA & \raggedright 8B & \raggedright\href{https://huggingface.co/mPLUG/GUI-Owl-1.5-8B-Instruct}{\nolinkurl{mPLUG/GUI-Owl-1.5-8B-Instruct}} & \citep{gui-owl-15} \\
\midrule
\raggedright OS-Atlas-Pro-7B & VLA & \raggedright 7B & \raggedright\href{https://huggingface.co/OS-Copilot/OS-Atlas-Pro-7B}{\nolinkurl{OS-Copilot/OS-Atlas-Pro-7B}} & \citep{wuatlas} \\
\raggedright ShowUI-2B & VLA & \raggedright 2B & \raggedright\href{https://huggingface.co/showlab/ShowUI-2B}{\nolinkurl{showlab/ShowUI-2B}} & \citep{lin2025showui} \\
\bottomrule
\end{tabular}
\endgroup
\end{table}

Model-specific prompting and output parsing are part of the evaluated configuration. Table~\ref{tab:model_interfaces} summarizes the parser and prompt template used for each baseline.

\begin{table}[!htbp]
\centering
\footnotesize
\caption{\textbf{Model interfaces.}
Output parser and prompt template used for each evaluated model.}
\label{tab:model_interfaces}
\begin{tabular}{@{}lll@{}}
\toprule
\textbf{Model} & \textbf{Parser} & \textbf{Prompt template} \\
\midrule
Gemma-4-26B-A4B-it        & qwen3         & standard \\
EvoCUA-8B                 & computer\_use & computer\_use \\
Qwen3-Omni-30B-A3B        & qwen3         & standard \\
Gemma-4-E4B-it            & qwen3         & standard \\
UI-TARS-1.5-7B            & uitars        & uitars \\
Qwen3-VL-8B               & qwen3         & standard \\
Qwen2.5-VL-7B             & standard      & standard \\
OpenCUA-7B                & opencua       & opencua \\
GUI-Owl-1.5-8B            & computer\_use & computer\_use \\
Gemma-3-12B-it            & standard      & standard \\
Qwen2.5-Omni-7B           & standard      & standard \\
OS-Atlas-Pro-7B           & osatlas       & osatlas \\
InternVL2.5-8B            & qwen3         & standard \\
ShowUI-2B                 & showui        & showui \\
\bottomrule
\end{tabular}
\end{table}

\subsection{Prompt Templates}
\label{app:prompts}

General-purpose VLMs use a shared prompt consisting of a system instruction defining the agent's role and a user template containing game metadata, available controls, interaction history, and the required action format. The complete VLM prompt is shown in Figure~\ref{fig:vlm_user_prompt}.

Specialized CUAs and VLAs retain their model-specific prompting and output formats, with only minor adaptations to the \playsuite interaction setting. The computer-use prompt is shown in Listing~\ref{lst:computer_use_prompt}; model-specific prompt and parser assignments are summarized in
Table~\ref{tab:model_interfaces}.

Prompting ablations involving explicit strategy traces and few-shot examples are reported in Appendix~\ref{app:prompt-results}.

\subsection{Execution details}
\label{app:execution}

\begin{sloppypar}
\noindent\textbf{PyWeek.}
For Python-based games, we instrument the game process directly by setting \texttt{SDL\_VIDEODRIVER=dummy} and overriding the internal display update cycle, allowing raw RGB observations to be captured directly from the game framebuffer. A virtual clock controls game progression and allows us to set the interaction cadence explicitly. We evaluate at interaction rates of 3~Hz and 0.33~Hz. Each run is capped at 10,000 game frames, corresponding to approximately 333 seconds ($\sim5.5$ minutes) of gameplay at 30~fps.
\end{sloppypar}

\noindent\textbf{itch.io.}
For standalone and browser-based games, we use Xvfb to provide an off-screen display for frame capture. Browser based games are controlled by freezing browser clocks, while native game processes are paused directly. We evaluate at 0.33~Hz, with each run spanning 300 seconds of game time and approximately 100 interaction steps. Inference pauses are removed from the judged video. We exclude 225 Construct~3 games whose game loops rely on Web Workers that cannot be reliably paused by our browser clock interception mechanism.

\noindent\textbf{Action interface.}
The shared action grammar supports single keys, chords, timed holds, mouse motion, left/right clicks, wheel events, and multiple clicks. Timed holds in the generalist VLM prompt are capped at two seconds.

\subsection{Video-LLM-as-a-judge additional details}
\label{app:judge}

\subsubsection{Judge configuration}
\label{app:judge_config}
All reported benchmark results use \texttt{Qwen/Qwen3.5-9B} as the Video-LLM judge. Gameplay videos are sampled at 1\,fps, with individual frames constrained to a maximum resolution of $512\times288$ pixels to balance visual detail with computational throughput. Decoding uses constrained JSON generation, temperature 0, and seed 41. The judge is served with batch size 8, up to 8 concurrent sequences, a 65,536-token context window, a 16,384-token generation limit, and up to 32,768 batched tokens. For \emph{itch.io}, the judge additionally receives game context that is separate from the information provided to the playing agent.

\subsubsection{Judge milestone prompt and progress rules}
\label{app:judge_prompt}
To ground the judge in each environment, the input prompt includes game-specific metadata. For \emph{itch.io}, this comprises the title, author, genre, description, community tags, and controls; for \emph{PyWeek}, it includes the title, author, genre, and project README. Figure~\ref{fig:judge_prompt} shows the milestone prompt and output schema; for readability, the game-specific context is replaced by a descriptive placeholder.

  \FloatBarrier
  \Needspace{9\baselineskip}%
  \begingroup
  \begin{mdframed}[backgroundcolor=gray!10,linecolor=gray!50,
    innerleftmargin=10pt,innerrightmargin=10pt,
    innertopmargin=10pt,innerbottommargin=10pt,
    splittopskip=10pt,splitbottomskip=10pt]
  \small\rmfamily\color{black}
  \setlength{\parindent}{0pt}\setlength{\parskip}{5pt}
  \textbf{IMPORTANT:}\par
\nopagebreak[4]
\begin{itemize}[nosep,leftmargin=15pt,topsep=2pt,partopsep=0pt,label={-}]
\item Return **ONLY valid JSON** matching the schema. No markdown fences, no prose.
\item Every field asks whether something is **visible at least once anywhere in the video**. Watching more footage can only ADD events, never remove them.
\item Base every answer strictly on visible evidence. When uncertain, answer false / 0 — never guess.
\end{itemize}

Task: Report which of the following concrete events are visible in this gameplay video. Do NOT rate quality, skill, or overall progress — only report observable events.\par

\textbf{Field definitions:}\par
\nopagebreak[4]
\begin{itemize}[nosep,leftmargin=15pt,topsep=2pt,partopsep=0pt,label={-}]
\item "menu\_or\_title\_seen": a menu, title screen, or splash screen is visible at any point.
\item "left\_menu\_into\_gameplay": the video shows a transition from a menu or title screen into live gameplay (the reverse does not count).
\item "player\_control\_evident": the game state visibly responds to player inputs at least twice (character moves purposefully, UI reacts to clicks/keys, selections change). Autonomous animation does not count.
\item "distinct\_interactions": integer count of DISTINCT visible interaction events (movement burst, click with UI response, attack, jump, item pickup...). Repeats of the identical action in place count once. Cap at 20. Estimate quickly — do NOT enumerate every interaction one by one; a rough count is fine.
\item "score\_or\_resource\_increased": a numeric score, resource count, currency, progress bar, or similar UI indicator visibly increases.
\item "new\_area\_or\_level\_reached": the player visibly enters an area, room, screen, or level not seen earlier in the video (scrolling within one screen does not count).
\item "checkpoint\_or\_objective\_completed": an explicit checkpoint, mission complete, level clear, or objective-done indicator appears.
\item "win\_or\_completion\_screen": an explicit victory / you-win / run-complete screen, or credits reached by finishing the game. A game-over, "you lose", death, or failure screen is NOT a completion screen — count those under "death\_or\_game\_over\_count" instead.
\item "death\_or\_game\_over\_count": integer count of visible deaths, game-over screens, or run-ending failures.
\item "stuck\_or\_looping": the majority of the video shows the same action or scene repeating with no visible change.
\item "milestones": one timestamped entry per TRUE boolean above (and per death), naming the field and what is visible on screen at that moment.
\end{itemize}

\begin{minipage}{\linewidth}

\textbf{Output JSON schema:}\par\nopagebreak[4]

\begin{lstlisting}[style=promptjson]

{
  "gameplay_happened": true|false,
  "menu_or_title_seen": true|false,
  "left_menu_into_gameplay": true|false,
  "player_control_evident": true|false,
  "distinct_interactions": number,
  "score_or_resource_increased": true|false,
  "new_area_or_level_reached": true|false,
  "checkpoint_or_objective_completed": true|false,
  "win_or_completion_screen": true|false,
  "death_or_game_over_count": number,
  "stuck_or_looping": true|false,
  "milestones": [
    {"t_sec": number, "field": string, "on_screen": string}
  ]
}

\end{lstlisting}

\end{minipage}

\textbf{\#\# Game Context}\par
\nopagebreak[4]
[Per-game context from the page snapshot or documentation.]\par

  \end{mdframed}%
  \nopagebreak[4]
  \captionsetup{hypcap=false}%
  \captionof{figure}{Milestone prompt; game context varies by video.}\label{fig:judge_prompt}%
  \endgroup
  \vspace{8pt}

Rather than directly assigning a progress score, the judge extracts observable gameplay milestones together with timestamped evidence. These include player control, score increases, entry into new areas or levels, completion of intermediate objectives, and game completion. The resulting structured output is mapped to the five progress levels using the fixed rules in Table~\ref{tab:progress_rule}. Rules are evaluated in descending priority and do not require lower-level conditions to hold.

\begin{table}[!htbp]
\centering
\small
\caption{\textbf{Deterministic mapping from milestones to progress.} Evaluate rows from top to bottom and return the first matching level. Conditions do not require the lower-level milestones to hold. Field names match the judge output.}
\label{tab:progress_rule}
\setlength{\tabcolsep}{5pt}
\renewcommand{\arraystretch}{1.3}
\begin{tabular}{@{}cp{.19\linewidth}p{.70\linewidth}@{}}
\toprule
\textbf{Score} & \textbf{Level} & \textbf{Condition}\\
\midrule
4 & Completed & \path{win_or_completion_screen}\\
3 & Substantial & \path{checkpoint_or_objective_completed}\newline
  \textbf{or} (\path{new_area_or_level_reached}\newline
  \textbf{and} \path{score_or_resource_increased})\\
2 & Partial & \path{new_area_or_level_reached}\newline
  \textbf{or} \path{score_or_resource_increased}\\
1 & Minimal & \path{player_control_evident}\newline
  \textbf{and} \path{distinct_interactions} $\geq 2$\\
0 & None & No condition above is satisfied.\\
\bottomrule
\end{tabular}
\end{table}

\subsubsection{Human and cross-judge validation}
\label{app:judge_validation}

We validate the judge on 60 gameplay videos, comprising 45 \emph{itch.io} and 15 \emph{PyWeek} runs spanning the evaluated model configurations and both history conditions. Six human annotators independently label each video while blinded to model and condition.

Human-only Krippendorff's $\alpha$ is 0.624, using squared distance between the five progress levels, and human annotators agree within one progress level in 93.0\% of pairwise comparisons. Qwen3.5-9B agrees within one level with individual human annotations in 90.4\% of comparisons; including it as a seventh rater yields $\alpha=0.605$.

To assess sensitivity to judge choice, we additionally evaluate Qwen3.5-27B, Gemma-4-31B-it, and Molmo2-8B on the same 60 videos. Table~\ref{tab:agreement_summary} shows that both Qwen configurations track human progress similarly, while the other judges exhibit greater variation.

\begin{table}[!htbp]
\centering
\small
\caption{\textbf{Judge--human agreement on the validation set.}
Agreement between each Video-LLM judge and the six human annotators,
averaged over available comparisons. Milestone agreement is computed
over the nine Boolean milestone fields.}
\label{tab:agreement_summary}
\begin{tabular}{@{}lrr@{}}
\toprule
\textbf{Judge} & \textbf{Milestone agreement} & \textbf{Within one} \\
\midrule
Qwen3.5-9B      & 84.4\% & 90.4\% \\
Qwen3.5-27B     & 83.9\% & 91.6\% \\
Gemma-4-31B-it  & 82.5\% & 77.3\% \\
Molmo2-8B       & 77.9\% & 81.2\% \\
\bottomrule
\end{tabular}
\end{table}

\subsection{Benchmark execution efficiency}
\label{app:execution_analyses}
\begin{figure}[ht]
\centering
\begin{subfigure}[t]{.325\textwidth}
\centering
\includegraphics[width=\linewidth]{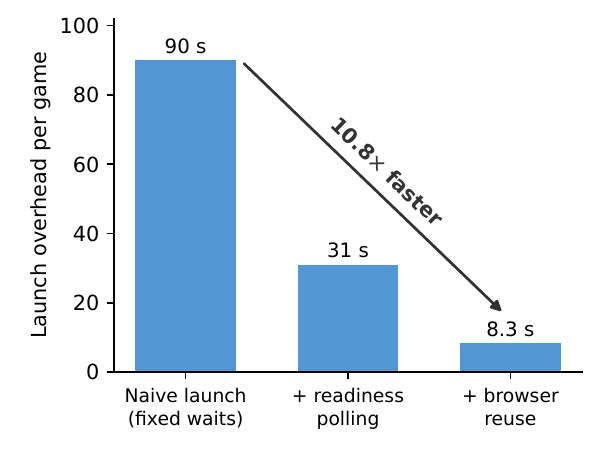}
\caption{{Launch overhead}}
\label{fig:execution-efficiency-a}
\end{subfigure}\hfill
\begin{subfigure}[t]{.325\textwidth}
\centering
\includegraphics[width=\linewidth]{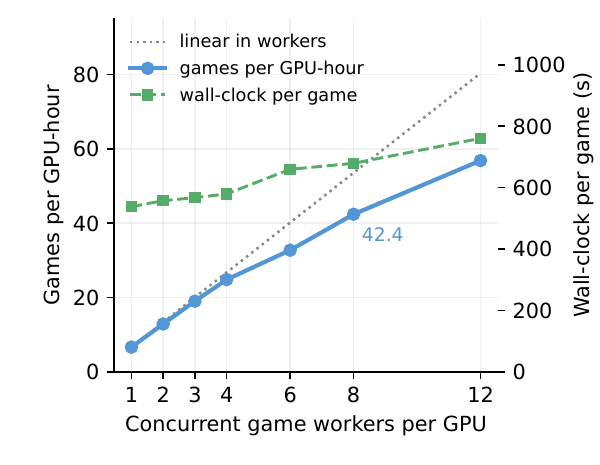}
\caption{{Concurrency on one H100}}
\label{fig:execution-efficiency-b}
\end{subfigure}\hfill
\begin{subfigure}[t]{.325\textwidth}
\centering
\includegraphics[width=\linewidth]{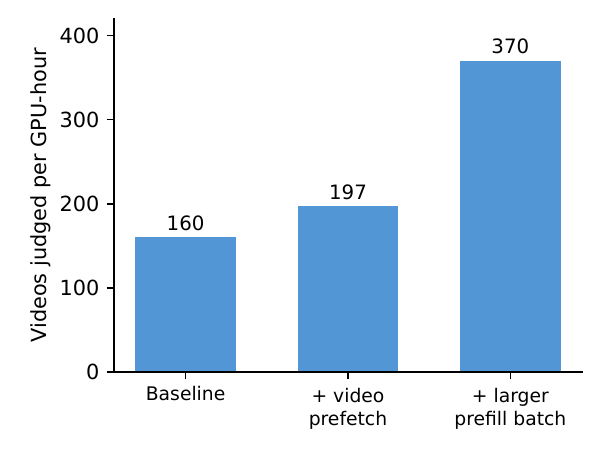}
\caption{{Judging throughput}}
\label{fig:execution-efficiency-c}
\end{subfigure}
\caption{{\textbf{Measured execution efficiency.} (a) Fixed overhead per game before the first captured frame, from 50-game launch tests. The naive launcher waits fixed intervals for the virtual display and the game page; readiness polling replaces these waits, and browser reuse keeps the browser and display running between games. (b) Qwen3-VL-8B on one H100, with concurrent game workers sharing a single vLLM server and the same games evaluated at each concurrency level. Throughput increases with worker count. (c) Increasing Qwen3.5-9B judge throughput through video prefetching and a larger batched-token budget (\texttt{max\_num\_batched\_tokens}=32,768) on 300s videos.}}
\label{fig:execution-efficiency}
\end{figure}

We characterize evaluation throughput using Qwen3-VL-8B on a single H100. The initial launcher relied on fixed waits for the virtual display, browser, and game to become ready. Replacing these waits with readiness checks allows execution to proceed as soon as each component is available, while keeping the browser and virtual display alive amortizes their initialization across games. Together, these changes reduce fixed launch overhead from approximately 90\,s to 8.3\,s per game. Sharing one vLLM server across concurrent game workers further increases throughput from 6.7 evaluations per GPU-hour with one worker to 42.4 with eight workers. Although throughput continues to increase beyond eight workers, higher concurrency increases CPU utilization and per-game wall-clock time, bringing some runs close to the harness safety timeout; we therefore use eight workers for the full benchmark. Figure~\ref{fig:execution-efficiency} summarizes these measurements.

We similarly optimize judging by overlapping video loading and decoding with model inference and by increasing the number of prompt tokens that vLLM can process in each batch. The latter is particularly important because each judge request contains tens of thousands of video tokens, making prompt prefill a substantial part of inference. On a controlled set of 300-s videos, these optimizations increase steady-state Qwen3.5-9B throughput from 160 to 370 videos per GPU-hour without changing the resulting milestone levels on comparable runs. When engine startup and other end-to-end overhead are included, the measured throughput is approximately 277 full-length \emph{itch.io} videos per GPU-hour.

\subsection{Additional benchmark analyses}
\label{app:additional_analyses}

\subsubsection{Performance across game genres}
\label{app:genre-results}
\begin{figure}[htbp]
\centering
\includegraphics[width=.98\textwidth]{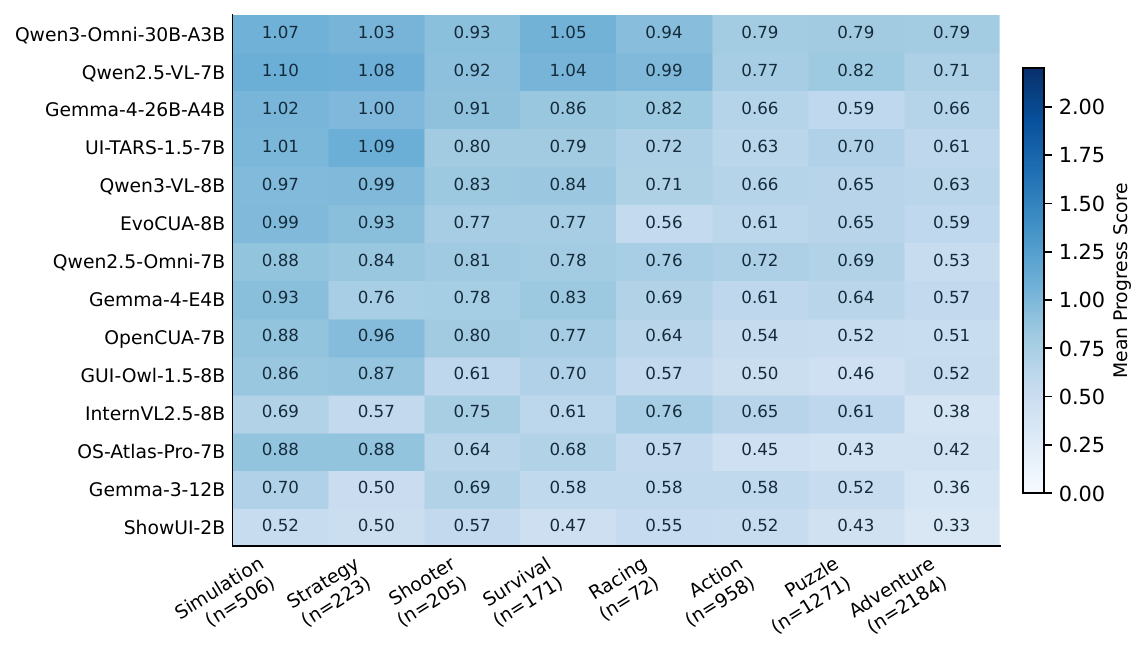}
\caption{\textbf{Per-model progress by genre on \emph{itch.io}.} Mean Progress Score in the no-history condition at 0.33\,Hz for each model and genre. Columns include genres with at least 60 games; $n$ denotes the number of games. Models are ordered by overall no-history performance.}
\label{fig:model_genre_itch}
\end{figure}

\begin{figure}[htbp]
\centering
\includegraphics[width=.98\textwidth]{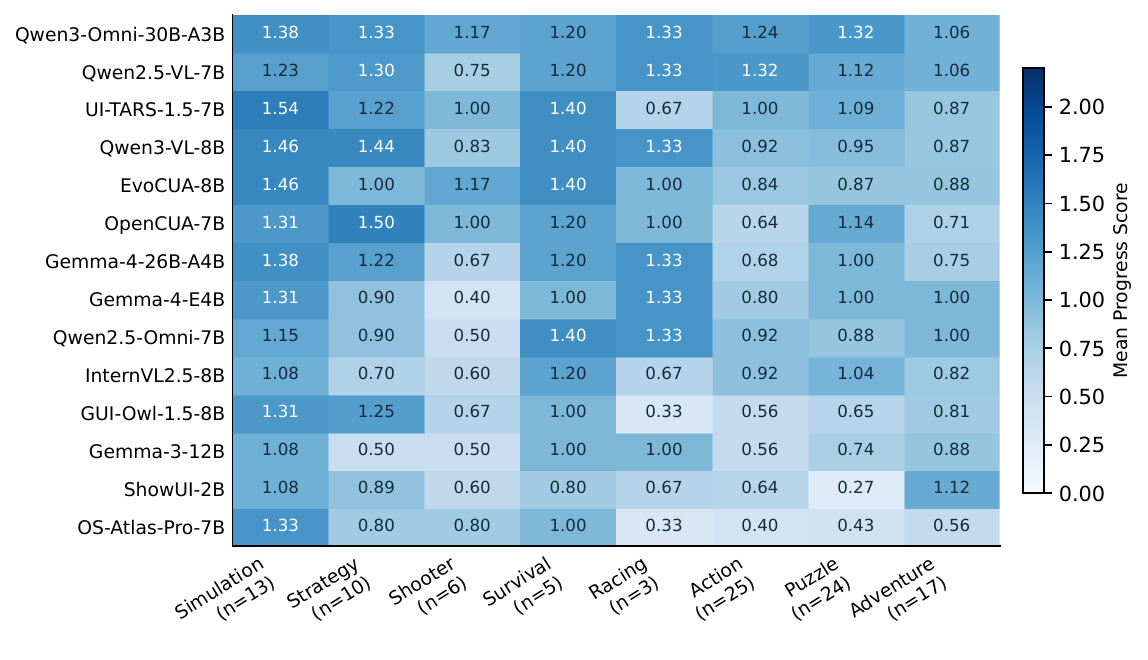}
\caption{\textbf{Per-model progress by genre on \emph{PyWeek}.} Mean Progress Score in the no-history condition at 0.33\,Hz for each model and genre. $n$ denotes the number of games. Models are ordered by overall no-history performance. Per-genre counts are small, so these results are primarily descriptive.}
\label{fig:model_genre_pyweek}
\end{figure}

Figures~\ref{fig:model_genre_itch} and~\ref{fig:model_genre_pyweek} show the per-model genre breakdowns in the no-history condition. On \emph{itch.io}, the aggregate pattern reported in the main text is broadly reflected across individual models, with higher progress on Simulation and Strategy and lower progress on Adventure and Puzzle. The \emph{PyWeek} results show greater variation across models, though this may partly reflect the small number of games in several genres. Table~\ref{tab:genre-history} additionally reports genre-level performance under both history conditions. Adding three interactions of history ($k=3$) reduces mean progress across every reported genre on both subsets, consistent with the overall history results.

\begin{table}[htbp]
\centering\small
\caption{\textbf{Genre-level progress under both history conditions.} Mean Progress Score for history lengths $k=0$ and $k=3$ on \emph{itch.io} and \emph{PyWeek}. $N$ denotes the number of games.}
\label{tab:genre-history}
\setlength{\tabcolsep}{5pt}
\begin{tabular}{@{}lrrrrrr@{}}
\toprule
& \multicolumn{3}{c}{\emph{itch.io}}
& \multicolumn{3}{c}{\emph{PyWeek}} \\
\cmidrule(lr){2-4}\cmidrule(lr){5-7}
Genre & $N$ & $k=0$ & $k=3$
      & $N$ & $k=0$ & $k=3$ \\
\midrule

Simulation & 506  & 0.893 & 0.800 & 13 & 1.293 & 1.173 \\
Strategy   & 223  & 0.858 & 0.704 & 10 & 1.054 & 0.702 \\
Shooter    & 205  & 0.771 & 0.682 &  6 & 0.763 & 0.720 \\
Survival   & 171  & 0.770 & 0.644 &  5 & 1.174 & 0.857 \\
Racing     & 72   & 0.705 & 0.654 &  3 & 0.976 & 0.714 \\
Action     & 958  & 0.620 & 0.591 & 25 & 0.817 & 0.699 \\
Puzzle     & 1271 & 0.606 & 0.546 & 24 & 0.893 & 0.771 \\
Adventure  & 2184 & 0.544 & 0.452 & 17 & 0.883 & 0.664 \\
\bottomrule
\end{tabular}
\end{table}

\subsubsection{Effect of action frequency}
\label{app:rate-results}
\begin{figure}[htbp]
\centering
\includegraphics[width=.98\textwidth]{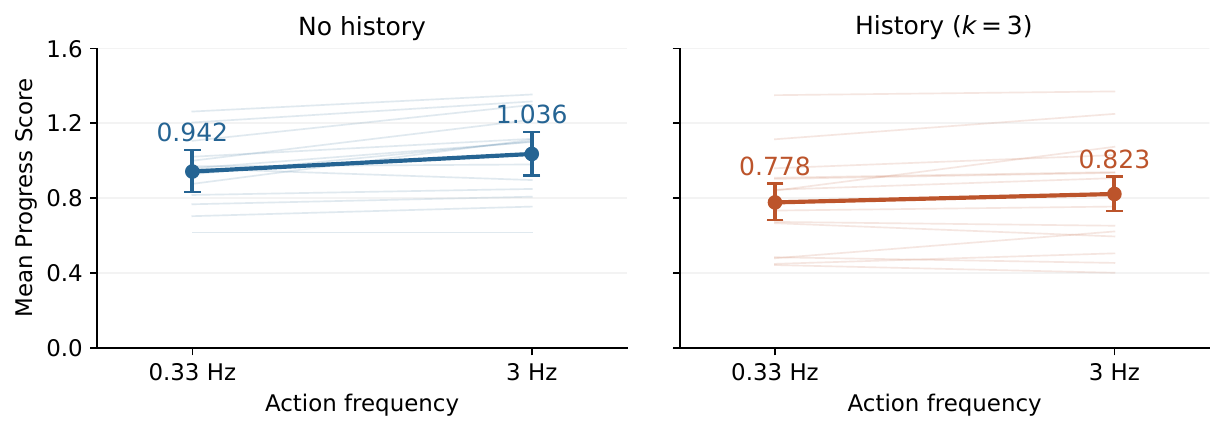}
\caption{\textbf{Effect of action frequency on \emph{PyWeek}.}
Mean Progress Score at 0.33\,Hz and 3\,Hz under the no-history and $k=3$ history conditions. Faint lines show individual model means and darker lines the mean over matched game--model pairs. Error bars denote 95\% game-bootstrap intervals.}
\label{fig:action_frequency}
\end{figure}

Figure~\ref{fig:action_frequency} shows the effect of increasing the \emph{PyWeek} interaction rate from 0.33\,Hz to 3\,Hz for each model. Mean progress increases in both conditions: from 0.942 to 1.036 without history and from 0.778 to 0.823 with $k=3$ history. The improvement is therefore 0.049 points larger without history.

Increasing the action rate does not provide clear evidence of an interaction with history. The mean gap between the no- history and history conditions increases from 0.165 points at 0.33\,Hz to 0.214 points at 3\,Hz, but a game-level bootstrap gives a 95\% interval of $[-0.027, 0.129]$ for this change. Moreover, per-model history effects at the two action rates are strongly correlated ($r=0.96$), indicating that models' relative sensitivity to history remains similar across interaction rates.

\subsubsection{Impact of prompting strategies}
\label{app:prompt-results}
\begin{figure}[htbp]
\centering
\includegraphics[width=.98\textwidth]{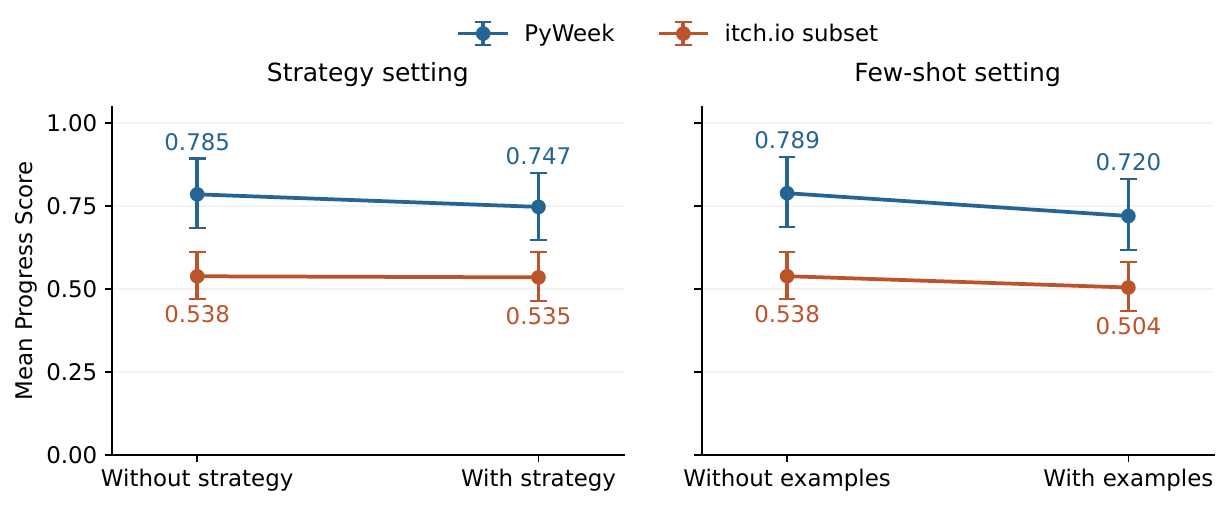}
\caption{\textbf{Effect of prompt additions.} Mean Progress Score with and without an explicit Strategy field (left) or three few-shot examples (right), for eight VLMs at $k=3$. Each comparison uses matched game--model cells. Error bars show 95\% game-bootstrap intervals for the means.}
\label{fig:prompting-comparison}
\end{figure}

We evaluate two modifications to the standard prompt for the eight VLMs: an explicit ``Strategy'' field requiring the model to state its overarching goal, and three few-shot examples demonstrating the expected response format. We restrict these ablations to the eight VLMs using the standard prompt template, since CUAs and VLAs rely on model-specific response formats and parsers that could confound prompt effects with parsing failures. Figure~\ref{fig:prompting-comparison} reports the resulting mean Progress Scores on \emph{PyWeek} and the 200-game \emph{itch.io} subset. Neither modification improves average performance. Adding an explicit strategy reduces mean progress from 0.785 to 0.747 on \emph{PyWeek}, while performance on \emph{itch.io} remains nearly unchanged (0.538 to 0.535). Few-shot examples reduce performance on both corpora, from 0.789 to 0.720 on \emph{PyWeek} and from 0.538 to 0.504 on \emph{itch.io}. Overall, adding prompt structure does not improve gameplay performance under our evaluation setup.

\subsubsection{Impact of temporal history}
\label{app:history}
\begin{figure}[htbp]
\centering
\includegraphics[width=.98\textwidth]{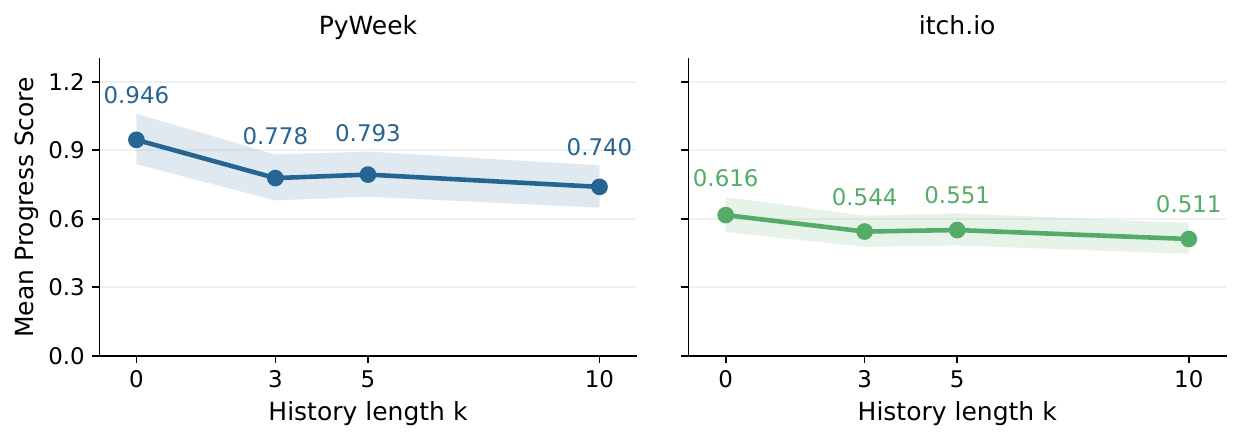}
\caption{\textbf{Impact of temporal history across both corpora.}
Mean Progress Score for history lengths $k\in\{0,3,5,10\}$ on \emph{PyWeek} and the 200-game \emph{itch.io} subset. No history achieves the highest mean performance on both corpora, while longer histories provide no consistent improvement.}
\label{fig:history-comparison}
\end{figure}

Figure~\ref{fig:history-comparison} reports performance across history lengths $k \in \{0,3,5,10\}$ for both subsets. On \emph{PyWeek}, mean Progress Score decreases from 0.946 at $k=0$ to 0.778 at $k=3$, remains similar at $k=5$ (0.793), and falls to 0.740 at $k=10$. The \emph{itch.io} subset shows the same pattern, decreasing from 0.616 at $k=0$ to 0.544 at $k=3$, 0.551 at $k=5$, and 0.511 at $k=10$. Thus, additional history does not improve average performance on either corpus, although individual models can differ in their response to temporal context.

\subsubsection{Cost--performance analysis}
\label{app:cost-results}
\begin{figure}[!b]
\centering
\includegraphics[width=.98\textwidth]{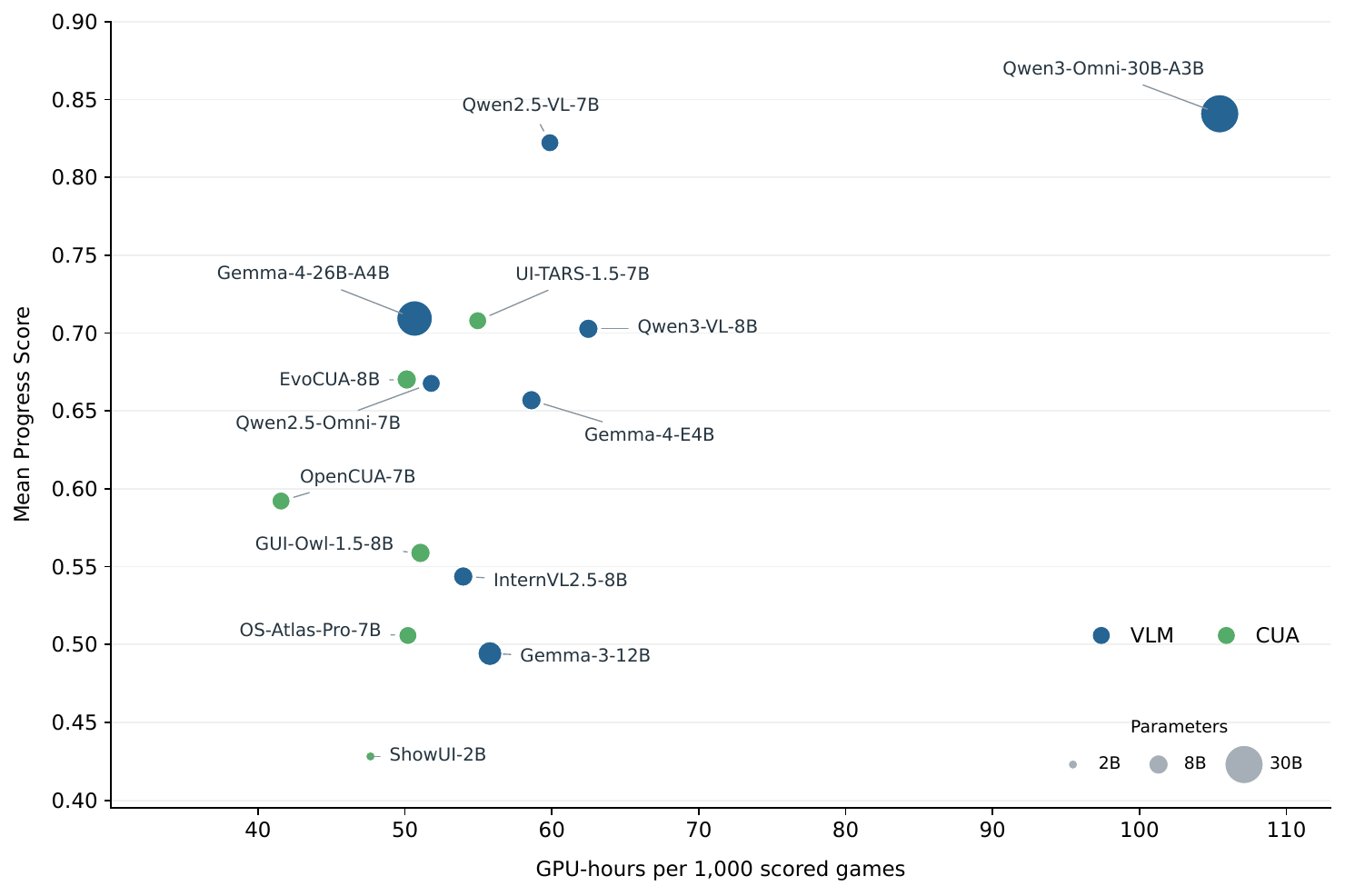}
\caption{\textbf{Cost--performance trade-off on \emph{itch.io}.} Mean Progress Score against H100 GPU-hours per 1,000 scored games in the no-history setting. Marker color denotes model family and marker size denotes parameter count.}
\label{fig:cost-performance}
\end{figure}

We measure realized evaluation cost for the no-history \emph{itch.io} campaign as allocated H100 GPU-hours per 1,000 scored games, including startup, idle allocation, retries, and failed or interrupted jobs. Judging cost is accounted for separately and excluded here. Figure~\ref{fig:cost-performance} shows substantial variation in both cost and performance across models. Qwen2.5-VL achieves a mean Progress Score of 0.822 at 59.9 GPU-hours per 1,000 games, while Qwen3-Omni reaches 0.841 at 105.5 GPU-hours. For Qwen3-Omni, both GPUs used per replica are included in the reported cost. Costs therefore reflect the realized compute required by each serving configuration.

\subsection{Failure Modes}

In this section we provide another example of the failure mechanisms discussed in the main text.

\begin{figure}[htbp]
    \centering
    \includegraphics[width=.8\textwidth]{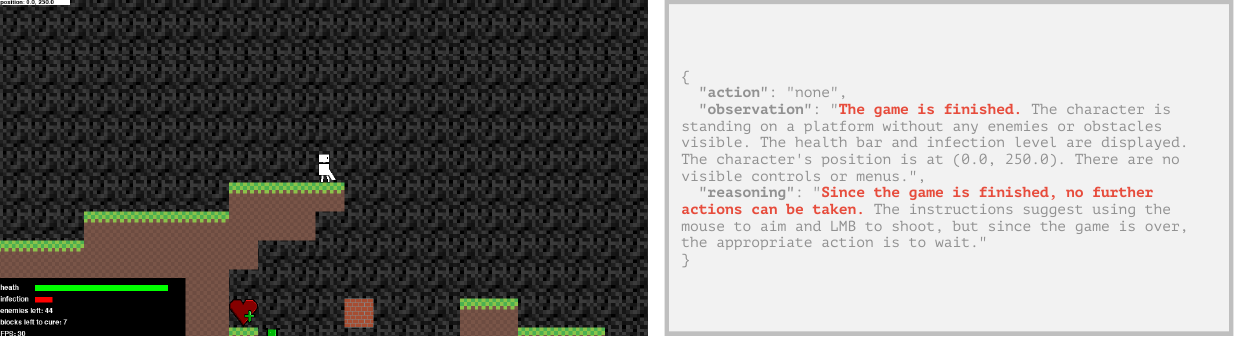}
    \caption{Example of a \emph{passive wait} failure mechanism. Qwen2.5-VL successfully starts the game, but since the initial game screen shows no visible enemies or apparent obstacles, it assumes the game has ended and waits, even if no on-screen information confirms that the game has been completed.}
    \label{fig:a-dying-world-failure-qwen}
\end{figure}

\subsection{Additional Tables}

%
  \FloatBarrier
  \Needspace{9\baselineskip}%
  \begingroup
  \begin{mdframed}[backgroundcolor=gray!10,linecolor=gray!50,
    innerleftmargin=10pt,innerrightmargin=10pt,
    innertopmargin=10pt,innerbottommargin=10pt,
    splittopskip=10pt,splitbottomskip=10pt]
  \small\rmfamily\color{black}
  \setlength{\parindent}{0pt}\setlength{\parskip}{5pt}
  You are an expert video game AI agent. Your goal is to play the provided game to maximize progress, score, or reach a win state.\par

\textbf{INSTRUCTIONS:}\par
\nopagebreak[4]
\begin{enumerate}[nosep,leftmargin=15pt,topsep=2pt,partopsep=0pt,start=1]
\item Analyze Controls: Read the "Game Description" and "Game Instructions". Ignore HTML tags like \textless{}br\textgreater{}.
\begin{itemize}[nosep,leftmargin=15pt,topsep=2pt,partopsep=0pt,label={-}]
\item Extract specific mappings (action: key). Keep this in memory but when prompt return only the key.
\item CRITICAL: If the instructions mentions "LMB" (Left Mouse Button), you must output a mouse action requiring coordinates ("CLICK: x, y").
\item If no controls are listed, infer them (e.g., Platformers usually use Arrows/WASD + Space).
\end{itemize}
\item Analyze Visuals: Look at the current frame. Identify the player character, enemies, obstacles, and the UI (health bar, score).
\item Detect Loops (Crucial): Compare the current frame to the history (if available).
\begin{itemize}[nosep,leftmargin=15pt,topsep=2pt,partopsep=0pt,label={-}]
\item If the visual state is identical to previous frames, you are STUCK.
\item If you are STUCK, you MUST change your strategy. Try a different key or direction.
\end{itemize}
\item Exploration: If the objective is unclear, interact with objects or move towards unexplored areas of the screen.
\end{enumerate}

  \end{mdframed}%
  \nopagebreak[4]
  \captionsetup{hypcap=false}%
  \captionof{figure}{Shared VLM system prompt.}\label{fig:system_prompt}%
  \endgroup
  \vspace{8pt}
  \FloatBarrier
  \Needspace{9\baselineskip}%
  \begingroup
  \begin{mdframed}[backgroundcolor=gray!10,linecolor=gray!50,
    innerleftmargin=10pt,innerrightmargin=10pt,
    innertopmargin=10pt,innerbottommargin=10pt,
    splittopskip=10pt,splitbottomskip=10pt]
  \small\rmfamily\color{black}
  \setlength{\parindent}{0pt}\setlength{\parskip}{5pt}
  You are playing an itch.io game called "\texttt{\textless{}\$game\_name\$\textgreater{}}".\texttt{\textless{}\$category\_suffix\$\textgreater{}}\par

\texttt{\textless{}\$game\_description\$\textgreater{}}\texttt{\textless{}\$readme\_section\$\textgreater{}}\par

\texttt{\textless{}\$history\_section\$\textgreater{}}\par

\textbf{=== CONTROL MAPPING RULES (STRICT) ===}\par
\nopagebreak[4]

\textbf{STEP 1: CHECK INSTRUCTIONS}\par
\nopagebreak[4]
\begin{itemize}[nosep,leftmargin=15pt,topsep=2pt,partopsep=0pt,label={-}]
\item The instructions above is the TRUTH.
\item If the instruction lists keys (e.g., "jump : w"), use EXACTLY those keys.
\item If the instruction lists "LMB" you must output "CLICK: x, y".
\end{itemize}

\textbf{STEP 2: FALLBACK DEFAULTS}\par
\nopagebreak[4]
\begin{itemize}[nosep,leftmargin=15pt,topsep=2pt,partopsep=0pt,label={-}]
\item ONLY use these if the instruction are missing or does not mention the action:
\begin{itemize}[nosep,leftmargin=15pt,topsep=2pt,partopsep=0pt,label={-}]
\item Movement: a (left), d (right), w (up), s (down)
\item Actions: space (jump/confirm), enter (start/confirm), escape (menu/back), click: x, y (shoot)
\item Other: z, x, c, q, e, r, f, 1, 2, 3, 4, 5
\end{itemize}
\end{itemize}

\textbf{=== HOW TO HANDLE MOUSE / LMB ===}\par
\nopagebreak[4]
\begin{itemize}[nosep,leftmargin=15pt,topsep=2pt,partopsep=0pt,label={-}]
\item MENUS: If you see text like "Start", "Start Game", "Play", "New Game", "Proceed" or similar output: CLICK: x, y (coordinates of the button).
\item LMB: If the instructions says "Shoot: LMB", look for an enemy or target on screen. Output: CLICK: x, y (coordinates of the enemy).
\item AIMING: If the instructions says "Aim with mouse", look at where you want to shoot and output: CLICK: x, y.
\end{itemize}

\textbf{=== HOW TO HANDLE GAME OVER / END STATES ===}\par
\nopagebreak[4]
\begin{itemize}[nosep,leftmargin=15pt,topsep=2pt,partopsep=0pt,label={-}]
\item If the frame shows a game-over, death, victory, or level-complete screen (e.g. text like "Game Over", "You Died", "Retry", "Resume", "Restart", "Continue", "Next Level", "Play Again", "Try Again"): CLICK: x, y (coordinates of the button).
\item If it is not clickable (no visible button, just text), use the fallback keys most likely to confirm/continue: enter, space, escape (try enter first unless the instructions say otherwise).
\end{itemize}

\textbf{=== ACTION FORMAT ===}\par
\nopagebreak[4]
\begin{itemize}[nosep,leftmargin=15pt,topsep=2pt,partopsep=0pt,label={-}]
\item Single key: space
\item Key chord (keys pressed together): shift+a
\item Timed hold (movement, charging): hold:left:0.8  or  hold:right+space:0.5  (seconds, max 2)
\item Wait: wait - take no action (use SPARINGLY)
\item Mouse click: CLICK: x, y
\item Right click (if the game uses it, e.g. RTS commands, alt fire): rightclick:x,y
\item Double click (open/activate, some UIs need it): click:x,y click:x,y
\item Mouse wheel (zoom / weapon switch / scroll lists): scroll:up or scroll:down (add :N for N notches, e.g. scroll:down:5)
\item Mouse move only (look around / aim, no click): move:x,y  or relative move:+dx,-dy (mouse-look games; smooth sweep, add :SECONDS for a slower pan)
\end{itemize}

\textbf{=== DECISION RULES ===}\par
\nopagebreak[4]
\begin{enumerate}[nosep,leftmargin=15pt,topsep=2pt,partopsep=0pt,start=1]
\item MENU SCREEN. If you see a game menu screenshot, click on text buttons ('Start', 'Start Game', 'Play', 'New Game', 'Proceed'): Use "CLICK: x, y".
\item GAME SCREEN:
\begin{itemize}[nosep,leftmargin=15pt,topsep=2pt,partopsep=0pt,label={-}]
\item Prefer keys defined in the instructions.
\item Try to move, jump, shoot, or interact based on what you see and on the instructions provided before.
\item If stuck (no changes in history), switch actions immediately (e.g., moving, jump, or shoot).
\end{itemize}
\item Loop Prevention: NEVER repeat the same action more than 2 times unless you see clear progress.
\end{enumerate}

You should think step by step and respond with the following format.\par

\textbf{Response:}\par
\nopagebreak[4]
Observation: ... (Describe what you see in the last frame, the character's current position, nearby obstacles and enemies, and if this is a menu or game.)\par
Reasoning: ... (Based on the current observation, think step by step and explain how you choose the action.)\par
Action: ... (One action line only: a key ("space"), a chord ("shift+a"), a timed hold ("hold:left:0.8"), "wait", or "CLICK: x, y". Do NOT add any other words.)\par

\textbf{Your response:}\par
\nopagebreak[4]

  \end{mdframed}%
  \nopagebreak[4]
  \captionsetup{hypcap=false}%
  \captionof{figure}{Shared VLM user prompt template, including game-context placeholders, action instructions, and the response format.}\label{fig:vlm_user_prompt}%
  \endgroup
  \vspace{8pt}
  \FloatBarrier
  \Needspace{9\baselineskip}%
  \begingroup
  \begin{mdframed}[backgroundcolor=gray!10,linecolor=gray!50,
    innerleftmargin=10pt,innerrightmargin=10pt,
    innertopmargin=10pt,innerbottommargin=10pt,
    splittopskip=10pt,splitbottomskip=10pt]
  \small\rmfamily\color{black}
  \setlength{\parindent}{0pt}\setlength{\parskip}{5pt}
  Please generate the next move according to the UI screenshot, instruction and previous actions.\par

Instruction: You are playing an itch.io game called "\texttt{\textless{}\$game\_name\$\textgreater{}}".\texttt{\textless{}\$category\_suffix\$\textgreater{}} Play the game to make as much progress as possible (start the game from any menu, then control the character).\par

\texttt{\textless{}\$game\_description\$\textgreater{}}\texttt{\textless{}\$readme\_section\$\textgreater{}}\par

\textbf{Game-specific guidance:}\par
\nopagebreak[4]
\begin{itemize}[nosep,leftmargin=15pt,topsep=2pt,partopsep=0pt,label={-}]
\item If the screen shows a menu, click the button that starts the game ("Start", "Play", "New Game" or similar) with left\_click.
\item During gameplay, PREFER keyboard actions via the `key` action: keys listed in the game instructions above take priority; otherwise try ["a"]/["d"]/["w"]/["s"] to move, ["space"] to jump or act, ["enter"] to confirm, ["escape"] for menus.
\item Use left\_click during gameplay only for aiming or shooting when the instructions mention the mouse.
\item If the screen shows a game-over, death, victory, or level-complete screen (text like "Game Over", "You Died", "Retry", "Resume", "Restart", "Continue", "Next Level", "Play Again", "Try Again"), left\_click the continue/retry button; if there is no visible button (just text), use the `key` action with ["enter"] first, then ["space"] or ["escape"].
\item Never use `terminate`; if nothing needs doing, use `wait`.
\item Do not repeat an action that produced no visible change; try a different key or click target.
\end{itemize}

\textbf{Previous actions:}\par
\nopagebreak[4]
\texttt{\textless{}\$history\_section\$\textgreater{}}\par

  \end{mdframed}%
  \nopagebreak[4]
  \captionsetup{hypcap=false}%
  \captionof{figure}{Computer-use template used by EvoCUA and GUI-Owl.}\label{lst:computer_use_prompt}%
  \endgroup
  \vspace{8pt}

\begin{figure}[t]
\begin{mdframed}[backgroundcolor=gray!10, linecolor=gray!50, innerleftmargin=10pt, innerrightmargin=10pt, innertopmargin=10pt, innerbottommargin=10pt]
\small

Examples:
\begin{itemize}[nosep, leftmargin=10pt, label={-}]
\item Observation: The screen displays the main menu of a game with "Start" and "Quit" options. There is no gameplay visible yet, indicating the game is not actively running. The player must begin the game to proceed. \\
Reasoning: Since this is a menu screen, the correct action is to click the "Start" button to begin the game. According to the control rules, for menu buttons like "Start", I must output the coordinates of the button. \\
Action: CLICK: 100, 500
\item Observation: The player character is standing in a dark, enclosed space, without visible enemies or obstacles yet. \\
Reasoning: Since the game has just started, the next logical step is to explore the area and interact with objects to progress. The instructions state that I need to use 'x' to interact with the environment. \\
Action: x
\item Observation: The player character is still at position (370.0, 700.0) on the same grassy platform. The environment has not changed, with dark matter and enemies visible on the right. The player is stuck and stagnant. \\
Reasoning: Repeated attempts to move right ('d') have not resulted in any progress, as the player remains in the same position. This indicates the action is ineffective. I must change strategy. I should try jumping. \\
Action: w

\end{itemize}

\vspace{\baselineskip}
Your response:
\end{mdframed}
\caption{Few-shot examples demonstrating structured reasoning and control-token selection for generalist VLM evaluation complementing the user prompt template shown in Figure~\ref{fig:vlm_user_prompt}.}
\label{fig:examples_in_prompt}
\end{figure}

\begin{figure}[!htbp]
\begin{mdframed}[backgroundcolor=gray!10, linecolor=gray!50,
    innerleftmargin=10pt, innerrightmargin=10pt,
    innertopmargin=10pt, innerbottommargin=10pt]
\small
\setlength{\parindent}{0pt}
\setlength{\parskip}{5pt}
You should think step by step and respond with the following format.

Response:

\textbf{Observation:} ... (Describe what you see in the last frame, the character's current position, nearby obstacles and enemies, and if this is a menu or game.)

\textbf{Strategy:} ... (Describe your overarching strategy at this point of the game.)

\textbf{Reasoning:} ... (Based on the current observation, think step by step and explain how you choose the action.)

\textbf{Action:} ... (One action line only: a key (\texttt{"space"}), a chord (\texttt{"shift+a"}), a timed hold (\texttt{"hold:left:0.8"}), \texttt{"wait"}, or \texttt{"CLICK: x, y"}. Do NOT add any other words.)

Your response:
\end{mdframed}
\caption{\textbf{Strategy setting.} Response-format excerpt for the eight VLMs using the shared prompt. The variant adds the \texttt{Strategy} field between \texttt{Observation} and \texttt{Reasoning} in the user template (Figure~\ref{fig:vlm_user_prompt}); the shared system instructions (Figure~\ref{fig:system_prompt}) are unchanged.}
\label{fig:strategy_prompt}
\end{figure}

\end{document}